\documentclass[preprint,12pt]{elsarticle}

\usepackage[cp1252]{inputenc}
\usepackage{graphicx}
\usepackage{algorithmic}
\usepackage{subfigure}
\usepackage{amssymb}
\usepackage{captcont}
\usepackage{array}
\usepackage{multirow}
\usepackage{hhline}
\usepackage{supertabular}
\usepackage{rotating}

\journal{Information Sciences}
\begin{document}
\begin{frontmatter}
\title{Ockham's Razor in Memetic Computing: Three Stage Optimal Memetic Exploration}

\author{Giovanni Iacca}
\ead{giovanni.iacca@jyu.fi}

\author{Ferrante Neri\corref{cor}}
\ead{ferrante.neri@jyu.fi}
\cortext[cor]{Corresponding author}

\author{Ernesto Mininno}
\ead{ernesto.mininno@jyu.fi}

\address{Department of Mathematical Information Technology,
           P.O.~Box~35~(Agora),
           40014 University of Jyv{\"a}skyl{\"a},
           Finland,
           Tel~+358-14-260-1211,
           Fax~+358-14-260-1021 }
           
\author{Yew-Soon Ong}
\ead{asysong@ntu.edu.sg}

\address{School of Computer Engineering
Nanyang Technological University
Block N4, 2b-39, Nanyang Avenue, Singapore 639798
Tel : +65-6790-6448
Fax: +65-6792-6559}

\author{Meng-Hiot Lim}
\ead{emhlim@ntu.edu.sg}
\address{School of Electrical and Electronic Engineering, Nanyang Technological University Singapore, S1-B1a-05, Nanyang Avenue, Singapore 639798 }

\begin{abstract}
Memetic Computing is a subject in computer science which considers complex structures as the combination of simple agents, memes, whose evolutionary interactions lead to intelligent structures capable of problem-solving. This paper focuses on Memetic Computing optimization algorithms and proposes a counter-tendency approach for algorithmic design. Research in the field tends to go in the direction of improving existing algorithms by combining different methods or through the formulation of more complicated structures. Contrary to this trend, we instead focus on simplicity, proposing a structurally simple algorithm with emphasis on processing only one solution at a time. The proposed algorithm, namely Three Stage Optimal Memetic Exploration, is composed of three memes; the first stochastic and with a long search radius, the second stochastic and with a moderate search radius and the third deterministic and with a short search radius.
The bottom-up combination of the three operators by means of a natural trial and error logic,  generates a robust and efficient optimizer, capable of competing with modern complex and computationally expensive algorithms. This is suggestive of the fact that complexity in algorithmic structures can be unnecessary, if not detrimental, and that simple bottom-up approaches are likely to be competitive is here invoked as an extension to Memetic Computing basing on the philosophical concept of Ockham's Razor. An extensive experimental setup on various test problems and one digital signal processing application is presented. 
Numerical results show that the proposed approach, despite its simplicity and low computational cost displays a very good performance on several problems,  and is competitive with sophisticated algorithms representing the-state-of-the-art in computational intelligence optimization.
\end{abstract}


\end{frontmatter}

\section{Introduction}
\label{intro}
Emerging technologies in computer science and engineering, as well as the demands of the market and the society, often impose the solution, in the every day life, of complex optimization problems. The complexity of today's problems is due to various reasons such as high non-linearities, high multi-modality, large scale, noisy fitness landscape, computationally expensive fitness functions, real-time demands, and limited hardware available (e.g. when the computational device is portable and cheap). In these cases, the use of exact methods is unsuitable because, in general, there is not sufficient prior knowledge  (hypotheses) on the optimization problem; thus,  computational intelligence approaches become not only advisable but often the only alternative to face the optimization. 

Scientific research in computational intelligence optimization can be classified into two general categories.
\begin{itemize}
\item In the first case, by following the No Free Lunch Theorem (NFLT) \cite{bib:Wolpert1997}, the application problem becomes the starting point for the algorithmic design, i.e. after an analysis of the problem, an algorithm containing components to address the specific features of the problem is implemented. Amongst domain specific algorithms, in 
 \cite{bib:Joshi1999} an ad-hoc Differential Evolution (DE) is implemented  for solving the multisensor fusion problem; in \cite{bib:Rogalsky2000}   DE based hybrid algorithm is designed to address an aerodynamic
design problem;  in \cite{bib:Fan2007}, an optimization approach is given with reference to the study of a material structure; in \cite{bib:Caponio2006} and \cite{bib:NeriMcDE2010} a computational intelligence approach is designed for a control engineering problem while in \cite{bib:Neri2007HIV1} and \cite{bib:Neri2006IEEEACM} a medical application for Human Immunodeficiency Virus (HIV) is addressed; in \cite{bib:Tirronen2009} a DE based hybrid  algorithm is implemented to design a digital filter for paper production industry. 
\item In the second case, computer scientists attempt to perform the algorithmic development with the aim of designing a robust algorithm, i.e. an algorithm capable to display a respectable performance on a diverse set of test problems. Usually,  the newly designed algorithms are tested on a set of test problems, see \cite{bib:Suganthan2005}. Some examples of articles containing this kind of approach are  \cite{bib:Weber2010SOCO}, \cite{bib:Weber2010GPEM}, \cite{bib:Mininno2011}, \cite{bib:Brest2006}, \cite{bib:Das2009}, and \cite{bib:Qin2009}.
\end{itemize}

Regardless of the aim of the designer, usually the algorithmic design does not result into a fully novel computational paradigm. On the contrary, computer scientists, on the basis of the results previously attained in literature perform an unexplored algorithmic coordination in order to  detect the lowest possible value of the objective functions. In our view, the most typical approaches which describe the ``mental  process''  of the computer scientists, when  a novel algorithmic design is performed, can be subdivided into the following three categories.       
\begin{enumerate}
\item Starting from an existing optimization algorithm, its structure is ``perturbed'' by slightly modifying the structure and adding on extra components.  Obviously, this approach attempts to obtain a certain performance improvement in correspondence to the proposed modifications.  A successful example of this research approaches is given in \cite{bib:Brest2006} where a controlled randomization on DE control parameters appear to offer a promising alternative to the standard DE framework, see also \cite{bib:Neri2010survey} . Other examples are given in \cite{bib:Das2009} and \cite{bib:Liang2006} where the variation operator combining the solutions of a population is modified in the context of DE and Particle Swarm Optimization (PSO), respectively. Other examples of PSO based algorithms obtained by modifying the original paradigm are shown in \cite{bib:Wu2010} and \cite{bib:Sun2011}. 
\item Starting from a set of algorithms, they are combined in a hybrid fashion with the trust that their combination and coordination leads to a flexible  structure  displaying a better performance than the various algorithms considered separately.   Two examples of recently proposed algorithms which are basically the combination, by means of activation probabilities, of various meta-heuristics are given in \cite{bib:Vrugt2009} and \cite{bib:Peng2010}.  A very similar concept is contained in the idea of ensemble, see \cite{bib:Mallipeddi2010} and \cite{bib:Mallipeddi2011}, where multiple strategy concur by means of a self-adaptive/randomized mechanism to the optimization of the same fitness function. Another similar concept is given in \cite{bib:Qin2009} where multiple search strategies, a complex randomized self-adaptation, and a learning mechanism are framed within a DE structure.  In \cite{bib:Nguyen2009} a combination of multiple algorithms is performed by assigning a certain success probability to each of them to detect the global optimum. In \cite{bib:Alba2005} and \cite{bib:Nguyen2009cellular} multiple algorithmic components are coordinated by means of the structural mapping of the population. In \cite{bib:Molina2010} a coordination scheme which promotes a sequence of local search activations is proposed.  In \cite{bib:Wang2009} a heuristic technique assists PSO in selecting the desired solutions while solving multi-objective optimization problems. Another good example of this algorithmic philosophy is the Frankenstein's PSO, see \cite{bib:deOca2009}, which combines several successful variants of PSO in order to make an ultimate PSO version. 
\item Starting from some knowledge of the problem features, the problem-knowledge is integrated within an algorithmic structure.  These algorithms usually make use of a theoretical background in order to enhance the performance of a metaheuristic framework. A typical case of this approach is in \cite{bib:Hansen2001} and \cite{bib:Hansen2003} where, on a solid theoretical basis, the search directions (by means of the distribution of solution) progressively adapt to the shape of the landscape.  This mechanism allows the algorithm to be rotational invariant and thus keen to handle the non-separability of the functions. By following a similar logic, two rotational invariant versions of DE are introduced in \cite{bib:Price2008}. In the context of PSO, a theoretical approach justifying the employment of inter-particle communication is presented in \cite{bib:Ghosh2011}.  
\end{enumerate} 

\begin{figure}
\includegraphics[width=.8\linewidth]{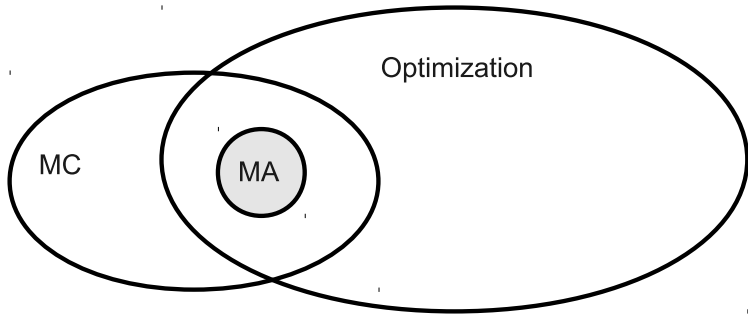}
\caption{Relationship between Memetic Algorithms, Memetic Computing, and Optimization}\label{MA_MC}
\end{figure}

According to the modern definition given in \cite{bib:OngMagazine2010}, these three categories fall within the umbrella name of Memetic Computing (MC). More specifically, MC is defined as ``a paradigm that uses the notion of meme(s) as units of information encoded in computational representations for the purpose of problem-solving'', where meme is an abstract concept which can be for example a strategy, an operator, or a search algorithm. In other words, a MC is strictly related to the concept of modularity and a MC structure can be seen as a collection of interactive modules whose interaction, in an evolutionary sense, leads to the generation of the solution of the problem. In this sense MC is a much broader concept with respect to a Memetic Algorithm (MA), which according to the definition in \cite{bib:Krasnogor2004a} is an optimization algorithm composed of an evolutionary framework and a list of local search algorithms activated within the generation cycle, of the external framework (see also \cite{bib:MoscatoNorman1989} and \cite{bib:Moscato1989}).  In this paper, we will refer to the unifying concept of MC and will consider each algorithm, in the light  of the definition in \cite{bib:OngMagazine2010}, as a composition of interacting and thus evolving modules (memes). More specifically, ``Memetic Computing is a broad subject which studies complex and dynamic computing structures composed of interacting modules (memes) whose evolution dynamics is inspired by the diffusion of ideas. Memes are simple strategies whose harmonic coordination allows the solution of various problems'', see \cite{bib:Neri2011Handbook}. Fig. \ref{MA_MC} further clarifies the relationship between MAs an MC.

Modern MC approaches for optimization aim at proposing an automatic evolution of the memes by using various kinds of feedback. The so-called meta-Lamarckian learning, see \cite{bib:Ong2004}, takes the feedback from the success of the algorithmic components belonging to the framework; self-adaptive and co-evolutionary approaches encode the memes within the solutions or in parallel populations and rely on their evolution, see \cite{bib:Krasnogor2005} and \cite{bib:Smith2007}; adaptive hyper-heuristic where the memes are coordinated by means of another algorithm, see e.g. \cite{bib:Cowling2000} and \cite{bib:Kononova2007}.  For a classification, see \cite{bib:Ong2006}.    
 
These modern approaches for optimization algorithms, regardless of the fact they are explicitly labelled as ``memetic'' or not, are very often (if not always) sophisticated and might be very complicated in the sense that can be characterized by many parts to be coordinated, by a large number of parameters, adaptive rules, and  various sampling distributions. The application of such kind of algorithms is usually infeasible on devices characterized by  limited hardware specificating resources, see \cite{bib:Mininno2008}, \cite{bib:Mininno2011}, \cite{bib:NeriMcDE2010}, and \cite{bib:NeriDEcDE2011}. In this paper, we discuss this research trend by posing the following research questions: ``Is the algorithmic complexity in optimization actually supported by the results? Is the complexity, in MC, an algorithmic feature which makes the algorithms any better?''. In order to address these questions, we propose an extremely simple optimization algorithm that is inspired by the principle of MC. The proposed algorithm, namely, Three Stage Optimal Memetic Exploration (3SOME), is a single solution algorithm which periodically performs three exploratory stages. During the long distance exploration, similar to a stochastic global search, a new solution is sampled within the entire decision space by using a crossover. During the middle distance exploration, a hyper-cube is generated around the candidate solution and some points are stochastically generated within it, in order to explore within a bounded region of the decision space. During the short distance exploration, a simple deterministic local search is applied to the solution, in order to quickly exploit the most promising search directions and refine the search. The coordination among the three stages of the search is performed by a minimalistic sequential trial-and-error rule.  Since this algorithm does not match the definition given in \cite{bib:Krasnogor2004a} (for example it is not population-based), we will not refer to it as MA. On the other hand, since the 3SOME algorithm performs an evolutionary combination (and coordination)  of memes (the exploration stages), we will refer to it as a MC approach.   
 
 In this paper we want to use the 3SOME algorithm as an example of a broader and deeper concept in MC: we want to propose instead of modifying and combining existing algorithms without a solid scientific background, a bottom-up approach which starts constructing the algorithm from  scratch and, most importantly, allows an understanding of functioning and potentials of each search operator composing the algorithm. Strictly connected to the bottom-up approach is the concept of simplicity in algorithmic design. With this statement we mean that, if the functioning of each component is understood in depth by the designer, few and simple components are enough to guarantee a high performance. Thus, the minimal complexity should be a goal of the algorithmic designer while unnecessary components and excessively complex coordination rules should be avoided. In other words, computational intelligence structures should be intelligent enough to make use of what is ``needed'' for the solution of the problem. The ``needs'' of the problem should be understood at first by the algorithmic designer and then properly encoded into the automatic coordination of the algorithmic components. This is, in our view, a cornerstone for the automatic design of the algorithms where instead of a human being this thinking line will be encoded within a machine. 
 
 The concept of minimal algorithmic complexity in MC is here presented as an extension to MC of the popular philosophical concept of the Ockham's Razor: the simplest explanation of natural phenomena is likely to be the closest to the truth. In an analogue way, an optimization problem can be seen as a natural phenomenon and the optimization algorithm should be based on its understanding (regardless this study is performed by a human being or a machine). This means that the optimization algorithm contains the countermeasures to handle the features of the fitness landscape. In order, on one hand, to avoid a waste of computational resources, and, on the other hand, to allow a proper algorithmic development when the fitness landscape changes (adding and removing memes/components/modules) only the strictly necessary components must be employed.        
 
The remainder of this paper is organized in the following way. Section~\ref{s:3SOME} describes the algorithmic details of 3SOME and explains its working principles and algorithmic philosophy.  Section~\ref{s:numres} shows  the experimental setup and numerical results of the present study by comparing the proposed 3SOME with base modern algorithms, MAs, and complex approaches. Section~\ref{s:conc} gives the conclusions of this
paper.

\section{Three Stage Optimal Memetic Exploration}\label{s:3SOME}
In order to clarify the notation used, we refer to the minimization problem of an objective function $f\left(x\right)$, where the candidate solution $x$ is a vector of $n$ design variables (or genes) in a decision space $D$. 

At the beginning of the optimization problem one candidate solution is randomly sampled within the decision space $D$. In analogy with compact optimization, see \cite{bib:Mininno2011}, \cite{bib:NeriMcDE2010}, and \cite{bib:NeriDEcDE2011}, we will refer to this candidate solution as elite and indicate to it with the symbol $x_e$. In addition to $x_e$, 3SOME makes use of another memory slot for attempting to detect other solutions. The latter solution, namely trial, is indicated with $x_t$. In the following sub-sections the three exploratory stages, separately, and then the coordination among them, are described in details.    

\subsection{Long distance exploration}

This exploration move attempts to detect a new promising solution within the entire decision space. While the elite $x_e$ is retained, at first, a trial solution $x_t$ is generated by randomly sampling a new set of $n$ genes. Subsequently, the exponential crossover in the fashion of DE is applied between $x_e$ and $x_t$, see \cite{bib:DEbook}. More specifically, one gene from $x_e$ is randomly selected. This gene replaces the corresponding gene within the trial solution $x_t$.  Then,  a set of random numbers between $0$ and $1$ are generated. As long as $rand\left(0,1\right)\leq Cr$, where the crossover rate $Cr$ is a predetermined parameter, the design variables from the elite $x_e$ are copied into the corresponding positions of the trial solution $x_t$. The first time that $rand\left(0,1\right)> Cr$, the copy process is interrupted. Thus, all the remaining design variables of the offspring are those initially sampled (belonging to the original $x_t$).  This exploration stage  performs the global stochastic search and thus attempts to detect unexplored promising basins of attraction. On the other hand, while this search mechanism extensively explores the decision space, it also promotes retention of a small section of the elite within the trial solution.  This kind of inheritance of some genes appears to be extremely beneficial in terms of performance with respect to a stochastic blind search (which would generate a completely new solution at each step). If the trial solution outperforms the elite, a replacement occurs. A replacement has been set also if the newly generated solution has the same performance of the elite. This is to prevent the search getting trapped in some plateaus of the decision space (regions of the decision space characterized by a null gradient). For the sake of clarity, the pseudo-code of the long distance exploration stage  is shown in Fig. \ref{long}

\begin{figure}
\begin{center}
\begin{scriptsize}
\fbox{\begin{minipage}[b]{120mm}
\begin{algorithmic}
    \STATE generate a random solution $x_t$ within $D$
    \STATE generate $i=round \left(n \cdot rand\left(0,1\right)\right)$
    \STATE $x_t[i]=x_e[i]$
    \WHILE {$rand\left(0,1\right)\leq Cr$}
    \STATE $x_t[i]=x_e[i]$
    \STATE $i=i+1$
    \IF {$i==n$}
    \STATE $i=1$
    \ENDIF
    \ENDWHILE
    \IF  {$f \left(x_t\right) \leq f \left( x_e\right)$}
    \STATE $x_e=x_t$
    \ENDIF
\end{algorithmic}
\end{minipage}
}
\end{scriptsize}
\end{center}
\caption{Long distance exploration} \label{long}
\end{figure}

It can easily be observed that for a given value of $Cr$, the meaning of the long distance exploration would change with the dimensionality of the problem. In other words, for low dimensionality problems the trial solution would inherit most of the genes from the elite while for  high dimensionality problems, only a small portion of $x_e$ would be copied into $x_t$.  In order to avoid this problem and make the crossover action independent on the dimensionality of the problem, the following quantity, namely inheritance factor, is fixed:    
\begin{equation}\label{alpha}
    \alpha_e \approx \frac{n_e}{n}
\end{equation}
where $n_e$ is the number of genes we expect to copy from $x_e$ into  $x_t$ in addition to that gene deterministically copied. The probability that $n_e$ genes are copied is $Cr^{n_e}=Cr^{n \alpha_e}$. In order to control the approximate amount of copied genes and to achieve that about $n_e$ genes are copied into the offspring we imposed that
\begin{equation}\label{Cralpha}
    Cr^{n\alpha_e}=0.5.
\end{equation}
It can easily be seen that, for a chosen $\alpha_e$, the crossover rate can be set on the basis of the dimensionality as follows:
\begin{equation}\label{crset}
Cr = \frac{1}{{\sqrt[{n{\alpha _e}}]{2}}}.
\end{equation}
In this way, we can choose the quantity of information that we expect it is inherited from $x_e$ to $x_t$ and calculate, on the basis of this and of the dimensionality of the problem, the corresponding crossover rate.  The long distance exploration is repeated until it does not detect a solution that outperforms the original elite. When a new promising solution is detected, and thus the elite is updated, the middle distance exploration is activated. Since the long distance exploration is supposed to perform the global search within the decision space, it is interrupted as soon as a potential promising area of the decision space is detected, in order to allow a more focused search around the new solution.   

\subsection{Middle distance exploration}
This exploration moves attempts to focus the search around promising solutions in order to determine whether the current elite deserves further computational budget or other unexplored areas of the decision space must be explored. A hyper-cube whose edge has side width equal to $\delta$ is constructed around the elite solution $x_e$.  Subsequently, for $k \times n$ times ($n$ is the dimensionality), one trial point $x_t$ is generated within the hypercube by random sampling and exponential crossover.  More specifically, by following the same mechanism described the long distance exploration stage, a random solution is generated and then a portion of the elite $x_e$ is copied into the randomly generated point.  Although the crossover mechanism is the same described above, the operation is performed by computing the crossover rate on the basis of $1-\alpha_e$. In practice, this means that most of the elite is copied into the trial solution and thus the middle distance exploration stage attempts to search along a limited number of search directions and make a randomized exploitation of the current elite solution.  The fitness of the newly generated point is then compared with the fitness of the elite. If the new point outperforms the elite (or has the same performance), $x_e$ is replaced by the new point, otherwise no replacement occurs. For the sake of clarity, the pseudo-code displaying the working principle of middle distance exploration is shown in Fig. \ref{middle}.

\begin{figure}
\begin{center}
\begin{scriptsize}
\fbox{\begin{minipage}[b]{120mm}
\begin{algorithmic}
\STATE construct a hypercube with side width $\delta$ centred in $x_e$
\FOR {$j=1:k \times n$}
\STATE generate a random solution $x_t$ within the hypercube
    \STATE generate $i=round \left(n \cdot rand\left(0,1\right)\right)$
    \STATE $x_t[i]=x_e[i]$
    \WHILE {$rand\left(0,1\right)\leq Cr'$}
    \STATE $x_t[i]=x_e[i]$
    \STATE $i=i+1$
    \IF {$i==n$}
    \STATE $i=1$
    \ENDIF
    \ENDWHILE
    \IF  {$f \left(x_t\right) \leq f \left( x_e\right)$}
    \STATE $x_e=x_t$
    \ENDIF
\ENDFOR
\end{algorithmic}
\end{minipage}
}
\end{scriptsize}
\end{center}
\caption{Middle distance exploration} \label{middle}
\end{figure}

After $k \times n$ comparison, if the elite has been updated a new hypercube with side width $\delta$ is constructed around the new elite and the search (by sampling $k \times n$ points) is repeated.  On the contrary, when the middle distance exploration does not lead to an improvement the short distance exploration is performed on $x_e$. Simply, the middle distance exploration is performed until it is successful. If the middle distance exploration does not lead to benefits, an alternative search logic (the deterministic logic of the short distance exploration) is applied. It is worthwhile commenting the setting of $k \times n$ points. Intuitively, multiple solutions are necessary in order to explore several coordinates. In addition, it is clear that the number of coordinates to explore should be related to the dimensionality of the problem. However, since diagonal movements are allowed, i.e. movements along a few coordinates simultaneously, the search  requires, on average, more than $n$ trial solutions. We empirically fixed the amount of trials to $k \times n$ since it showed a good performance for various dimensionality values. 

\subsection{Short distance exploration}

This exploration move attempts to fully exploit promising search directions. The meaning of this exploration stage is to perform the descent of promising basins of attraction and possibly finalize the search if the basin of attraction is globally optimal. De facto, the short distance exploration is a simple steepest descent deterministic local search algorithm, with an exploratory move similar to that of Hooke-Jeeves algorithm, see \cite{bib:Hooke-Jeeves1961}, or the first local search algorithm of the multiple trajectory search, see \cite{bib:Tseng2008}. The short distance exploration stage requires an additional memory slot, which will be referred to as $x_s$ ($s$ stands for short).  Starting from the elite $x_e$, this local search, explores each coordinate $i$ (each gene) and samples $x_s[i]=x_e[i]-\rho$, where $\rho$ is the exploratory radius. Subsequently, if $x_s$ outperforms $x_e$, the trial solution $x_t$ is updated (it takes the value of $x_s$), otherwise a half step in the opposite direction $x_s[i]=x_e[i]+\frac{\rho}{2}$ is performed.  Again, $x_s$ replaces $x_t$ if it outperforms $x_e$. If there is no update, i.e. the exploration was unsuccessful, the radius $\rho$ is halved. This exploration is repeated for all the design variables and stopped when a prefixed budget (equal to 150 iterations) is exceeded.  For the sake of clarity, the pseudo-code displaying the working principles of the short distance exploration is given Fig. \ref{short}. 

\begin{figure}
\begin{center}
\begin{scriptsize}
\fbox{\begin{minipage}[b]{120mm}
\begin{algorithmic}
\WHILE{local budget condition}
\STATE $x_t=x_e$
\STATE $x_s=x_e$
\FOR{$i=1:n$}
\STATE $x_s[i]=x_e[i]-\rho$
\IF{$f \left(x_s\right) \leq f \left(x_t\right)$}
\STATE $x_t=x_s$
\ELSE
\STATE $x_s[i]=x_e[i]+\frac{\rho}{2}$
\IF{$f \left(x_s\right) \leq f \left(x_t\right)$}
\STATE $x_t=x_s$
\ENDIF
\ENDIF
\ENDFOR
\IF{$f \left(x_t\right) \leq f \left(x_e\right)$} 
\STATE $x_e=x_t$ 
\ELSE 
\STATE $\rho=\frac {\rho}{2}$
\ENDIF
\ENDWHILE
\end{algorithmic}
\end{minipage}
}
\end{scriptsize}
\end{center}
\caption{Short distance exploration} \label{short}
\end{figure}

After the application of the short distance exploration, if there is an improvement in the quality of the solution, the focused search of middle distance exploration is repeated subsequently. Nevertheless, if no improvement in solution quality is found, the long distance search is activated to attempt to find new basins of attractions. 

\subsection{Coordination of the exploration stages}
The proposed 3SOME is a three stage algorithm where three operators, in a memetic fashion, compete and cooperate, see \cite{bib:Krasnogor2004b}, in order to perform the global optimization search.  At the beginning of the optimization process, the elite solution $x_e$ is sampled in the decision space $D$. The long distance exploration is performed until a new solution outperforming the original elite is generated. The main idea is that the global search is the most explorative component and therefore it should continue until promising areas of the decision space is detected. Then, on the newly generated promising solution, the search is focused on a smaller portion of the decision space (on the interesting region under observation) and so the middle distance exploration is performed. This second stage exploration is continued until its application is successful. In other words, the middle distance exploration performs the search and moves within the decision space by following the most promising search directions. When this exploration is no longer capable of improving upon the elite, the search is focused with an alternative search logic, i.e. the short distance exploration is performed. The third stage of exploration is important since it compensates, with its deterministic rules, the randomness of the other two stages and eventually exploits the solutions when the first two exploration stages do not succeed at improving upon the elite solution. After the short distance exploration, if it fails, the new elite $x_e$ undergoes again the long distance exploration in search for new promising basins of attractions. On the contrary, if the search turned out to be successful, the newly updated elite undergoes middle distance exploration again in order to continue the search in the neighbourhood of this solution, but by means of the deterministic search logic of the short distance exploration. The search is interrupted when the global budget is reached. For the sake of clarity, the pseudo-code displaying the working principle of 3SOME and highlighting the coordination amongst the three levels of exploration is given in Fig. \ref{coordination}.  The choice of the acronym 3SOME is due to a parallelism with a three player golf match where two players cooperate with each other and compete with a third player. In our case, the long distance exploration tries to ``shoot'' and approach the green, the middle distance exploration jointly with the short distance exploration attempt to move towards the ``hole'' and end the game. If the latter two memes fail at improving upon the solution, the long distance exploration will ``play'' again attempting to detect a new promising area.

\begin{figure}
\begin{center}
\begin{scriptsize}
\fbox{\begin{minipage}[b]{120mm}
\begin{algorithmic}
\STATE generate the solution $x_e$
\WHILE{global budget condition}      
\WHILE{$x_e$ is not updated}
\STATE apply to $x_e$ the long distance exploration as in Fig. \ref{long}
\ENDWHILE
\WHILE{$x_e$ is updated}
\STATE apply to $x_e$ the middle distance exploration as in Fig. \ref{middle}
\ENDWHILE
\STATE apply to $x_e$ the short distance exploration as in Fig. \ref{short}
\IF {$x_e$ has been updated}
\STATE apply middle distance exploration as in Fig. \ref{middle}
\ELSE
\STATE apply long distance exploration as in Fig. \ref{long}
\ENDIF
\ENDWHILE
\end{algorithmic}
\end{minipage}
}
\end{scriptsize}
\end{center}
\caption{Coordination of the exploration stages} \label{coordination}
\end{figure}

As a final remark, a toroidal management of the bounds has been implemented. This means that if, along the dimension $i$, the design variable $x[i]$ exceeds the bounds of a value $\zeta$, it is reinserted from the other end of the interval at a distance $\zeta$ from the edge, i.e. given an interval $\left[a,b\right]$, if $x[i]=b+\zeta$ it takes the value of $a+\zeta$.

\subsection{Algorithmic philosophy: Ockham Razor in Memetic Computing}

As previouslt stated the proposed 3SOME is technically not a MA since it is not a population-based algorithm composed of an evolutionary framework with one or more local searchers. On the contrary, 3SOME is a simple single-solution algorithm which combines sequentially and by means of a minimalistic trial and error  three-operator logic. The interactions of these simple operators allow the algorithmic coordination towards the search of the global optimum. As mentioned in the self-explicative names, the long distance exploration has a global search role, the middle distance exploration attempts to focus the search and, as a magnifier, moves and better analyse interesting  areas of the decision space while the short distance exploration performs a fine-tuning of the most promising genotypes. On the other hand, the search is not interrupted with the fine tuning but if the last search step was successful, the focused search is continued by means of the middle distance exploration (and thus with a complementary search logic) in order to better exploit the promising areas and slightly move the search area if necessary. When the deterministic fine tuning completely exploited the area, i.e. the application of the short distance exploration does not lead to any benefit, the global search operator for searching for new areas of the decision space is activated again. 

A first consideration on this algorithmic structure could be done regarding the balance between exploration and exploitation. In the context of MAs, or more generally MC, this balance is often seen as the coordination between global and local search, see \cite{bib:Ishibuchi2004}.  In the case of the 3SOME algorithm, three levels of exploration are proposed in order to perform the global optimization. It is important to remark that, according to our preliminary experiments, if one of the three exploration levels is removed, the algorithmic performance drops dramatically and the algorithm becomes incapable to handle  most of the fitness landscapes.  This fact appears to be in agreement with the concept of robust MA, see \cite{bib:Krasnogor2004b}, where the employment of multiple search operators having different features  is proposed. Multiple and diverse operators are then supposed to compensate and complete each other in order to solve sets of diverse problems. This approach is also similar to the employment of two local search algorithms, one containing a certain degree of randomization, the other being fully deterministic, as proposed in \cite{bib:Caponio2006}. In this light, the employment of a middle distance randomized search and a short distance fully deterministic search is not a novel concept. 

\begin{figure}
\includegraphics[width=.8\linewidth]{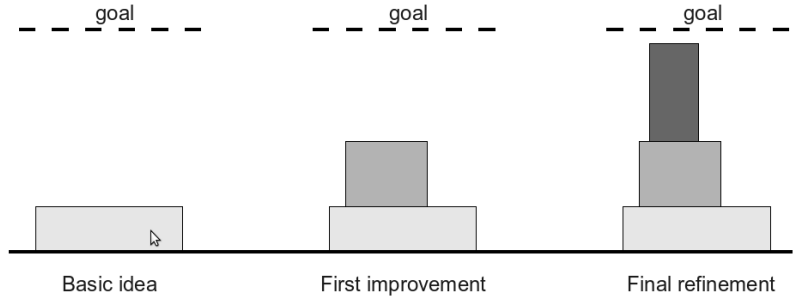}
\caption{Bottom-up design of an algorithm}\label{bottom-up}
\end{figure}

\begin{figure}
\includegraphics[width=.8\linewidth]{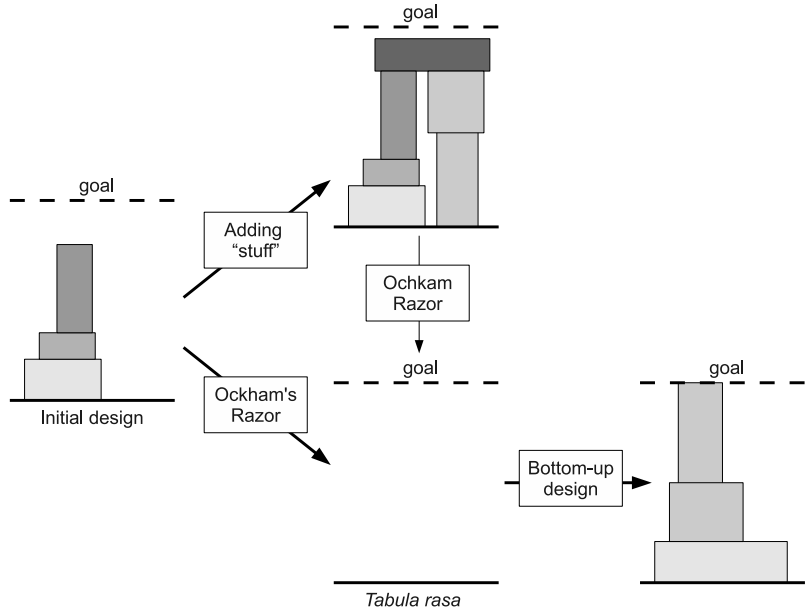}
\caption{Our view of current state-of-the-art in MC optimization and our proposal}\label{occamfig}
\end{figure}

On the other hand, the low level implementation which makes use of minimalistic components (memes) instead of combining existing complex algorithms is a radically novel idea in MC. In other words, as stated above, in MC often algorithms are designed either by adding a local search within a successful algorithm in order to enhance its performance or  by building up a patchwork of several successful algorithms in order to generate a hybrid structure which should outperform each of its component. In this case, we are proposing a bottom-up approach which considers as memes, four simple operators: random search, deterministic search, inheritance of a part of the solution, and restriction of the search space. It must be observed that the proposed 3SOME is a combination of these four ingredients within the three exploration algorithms. The memes are added one-by-one until the performance goal of making a competitive algorithm is reached. As an obvious consequence, the proposed resulting algorithm is much simpler than most of the MC approaches proposed during the last decade. However, as it will be shown in Section \ref{s:numres}, 3SOME does not perform worse than modern complex algorithms. In our view, for a correct development of the subject, it is crucial to be able to understand  the role,  potential, and  features, of each search operator. Such understanding will allow, in the long run, the automatic algorithmic design which is considered the next level of MC, see \cite{bib:Meuth2009}. A graphical representation of the bottom-up approach is given in Fig. \ref{bottom-up}, while our view on the current algorithmic design tendency in the field and a summary of the proposed Ockham's razor is illustrated  in Fig. \ref{occamfig}.

With the 3SOME algorithm, we want to claim that in order to obtain an algorithm with a  good performance, a correct approach in MC is to start from a \emph{tabula rasa} (blank slate, white paper) and build up the algorithm with a few simple memes, rather than starting from fully working algorithms and complicate them further. In any case, the complexity of the algorithm should never be excessive with respect to the problem(s) to be solved. In this sense we would like to retrieve a classical philosophical concept, the Ockham's Razor (from the medieval philosopher William of Ockham), also known as \emph{lex parsimoniae} (law of parsimony) and extend it to MC. Ockham's Razor is expressed in the following way: \emph{entia non sunt multiplicanda praeter necessitatem}  (entities must not be multiplied beyond necessity). In MC, entities are the memes, in the sense of search operators, and their quantity refers to their multiple coordination to construct complex algorithmic structures.  

Finally, it must be highlighted that 3SOME requires only three memory slots, i.e. $x_e$, $x_t$, and $x_s$, and thanks to its simple structure (few programming lines) can be easily implemented in a low level programming language  (e.g. to program a Graphic Processing Unit by using multiple parallel and concurrent 3SOME instances) and into devices characterized by a limited hardware, such as embedded systems, or simple mobiles, see \cite{bib:NeriMcDE2010} and \cite{bib:NeriDEcDE2011}. In addition, due to its simple structure,  the proposed 3SOME is characterized by a modest computational overhead with respect to modern MC approaches. It can be immediately observed that since 3SOME does not make use of computationally expensive structures such as an archive, a learning process, or a database of enhancements, see e.g. \cite{bib:Zhang2009} and \cite{bib:Qin2009}, it has a very modest execution time for a given amount of fitness evaluations.   




\section{Numerical Results}\label{s:numres}
The $30$ test problems in \ref{test_desc} have been considered in this study.

The $30$ selected problems are diverse in terms of modality, separability and dimensionality features. They have been selected in order to offer a more varied test bed. It must be observed that the problems $f_{24}-f_{30}$ compose the entire test-bed for large scale optimization competition, proposed during the IEEE Congress on Evolutionary Computation in 2008. 

All the algorithms in this paper have been run for the above-mentioned test problems.  For each algorithm, $30$ independent runs have been performed. The budget of each single run has been fixed to $5000 \cdot n$ fitness evaluations. 

In order to solve the above-mentioned optimization problem, 3SOME has been run by setting the inheritance factor $\alpha_e=0.05$, the width $\delta$ of the hypercube for middle distance exploration equal to $20 \%$ of the total decision space width, coefficient of generated points $k=4$, and the initial exploratory radius $\rho$ for short distance exploration equal to $40 \%$ of the total decision space width.   

The proposed 3SOME algorithm has been compared with nine modern algorithms. We grouped them into the following three categories:

\begin{enumerate}
\item Modern baseline algorithms
\item MC approaches recently proposed in literature
\item Efficient optimization algorithms, composed of multiple components and representing the-state-of-the-art in optimization    
\end{enumerate}

The following subsections describe the numerical comparisons for each of the category.

\subsection{Comparison with modern baseline algorithms}
The first category is represented by the following algorithms:

\begin{itemize}
\item Estimation of Distribution Algorithm with MultiVariate Gaussian model (EDA$_{mvg}$) proposed in \cite{bib:Yuan2005}. EDA$_{mvg}$ has been run with learning rate $\alpha=0.2$, population size $N_p=50$, selection ratio $\tau=0.3$, and maximum amplification value $Q=1.5$.
  \item (1+1) Covariance Matrix Adaptation Evolution Strategy ((1+1)-CMA-ES), proposed in \cite{bib:Igel2006}, with the parameter $\sigma=0.5$.
  \item 2-Opt based Differential Evolution (2OptDE) proposed in \cite{bib:Chiang2010} has been run with population size  $N_p=50$, scale factor $F=0.5$ and crossover rate $Cr=0.1$. The 2OptE algorithm has been run with the 2-Opt/2 mutation strategy since in \cite{bib:Chiang2010} it appeared to display the best performance amongst the  considered mutation strategies. 
\end{itemize} 

Table \ref{rescomp1} shows the average of the final results (after bias removal when the bias is present) detected by each algorithm $\pm$ the corresponding standard deviation values calculated over the  $30$ runs. The best results are highlighted in bold face. In order to strengthen the statistical significance of the results, the Wilcoxon Rank-Sum test has also been applied according to the description given in \cite{bib:Wilcoxon1945}, where the confidence level has been fixed at 0.95. Table \ref{Wilcoxoncomp1} shows the results of the Wilcoxon test for 3SOME against the other algorithms considered in this study. A ``+'' indicates the case in which 3SOME statistically outperforms the algorithm labelled on the top of the column for the corresponding test problem; a ``='' indicates that a pairwise comparison leads to success of the Wilcoxon Rank-Sum test, i.e., the two algorithms have the same performance; a ``-'' indicates that 3SOME is outperformed.

\begin{table}[t]
\caption{Fitness comparison with baseline algorithms} \label{rescomp1}
\begin{center} 
\begin{tiny}
\begin{tabular}{c|c|c|c|c} 
\hline 
\hline  
\multicolumn{1}{l|}{Test Problem} & EDA$_{mvg}$ & (1+1)-CMA-ES & 2OptDE & 3SOME \\ \hline  
\multicolumn{ 1}{c|}{$f_{1}$}  & 6.955e+01 $\pm$ 1.47e+02 & 1.961e-27 $\pm$ 1.47e-27 & 1.414e-08 $\pm$ 1.52e-08 & \textbf{0.000e+00} $\pm$ \textbf{0.00e+00} \\ \hline  
\multicolumn{ 1}{c|}{$f_{2}$}  & 3.145e+02 $\pm$ 3.01e+02 & \textbf{6.430e-26} $\pm$ \textbf{7.81e-26} & 1.428e+02 $\pm$ 1.21e+02 & 1.604e-23 $\pm$ 2.99e-23 \\ \hline  
\multicolumn{ 1}{c|}{$f_{3}$}  & 2.319e+06 $\pm$ 5.55e+06 & \textbf{1.017e+00} $\pm$ \textbf{1.80e+00} & 1.937e+01 $\pm$ 1.27e+00 & 5.393e+01 $\pm$ 1.60e+02 \\ \hline  
\multicolumn{ 1}{c|}{$f_{4}$}  & 3.271e+00 $\pm$ 4.71e-01 & 1.946e+01 $\pm$ 1.77e-01 & 3.740e-05 $\pm$ 1.45e-05 & \textbf{4.796e-14} $\pm$ \textbf{1.21e-14} \\ \hline  
\multicolumn{ 1}{c|}{$f_{5}$}  & 3.406e+00 $\pm$ 1.07e+00 & 1.942e+01 $\pm$ 1.99e-01 & \textbf{4.086e-05} $\pm$ \textbf{1.63e-05} & 1.645e+01 $\pm$ 6.39e+00 \\ \hline  
\multicolumn{ 1}{c|}{$f_{6}$}  & 2.638e+02 $\pm$ 3.90e+01 & 9.842e-03 $\pm$ 1.05e-02 & 1.557e+00 $\pm$ 2.60e-01 & \textbf{4.313e-03} $\pm$ \textbf{4.55e-03} \\ \hline  
\multicolumn{ 1}{c|}{$f_{7}$}  & 2.735e+02 $\pm$ 4.50e+01 & 2.843e-01 $\pm$ 2.34e-01 & 1.531e+00 $\pm$ 3.10e-01 & \textbf{1.788e-01} $\pm$ \textbf{1.89e-01} \\ \hline  
\multicolumn{ 1}{c|}{$f_{8}$}  & 1.770e+02 $\pm$ 1.35e+01 & 1.912e+02 $\pm$ 3.13e+01 & 2.246e+02 $\pm$ 1.33e+01 & \textbf{2.487e-13} $\pm$ \textbf{1.16e-13} \\ \hline  
\multicolumn{ 1}{c|}{$f_{9}$}  & \textbf{1.816e+02} $\pm$ \textbf{1.26e+01} & 1.892e+02 $\pm$ 4.33e+01 & 2.354e+02 $\pm$ 1.75e+01 & 2.281e+02 $\pm$ 4.76e+01 \\ \hline  
\multicolumn{ 1}{c|}{$f_{10}$}  & 1.402e+04 $\pm$ 5.35e+04 & 3.980e+02 $\pm$ 1.14e+02 & 2.349e+02 $\pm$ 1.23e+01 & \textbf{3.661e+01} $\pm$ \textbf{8.03e+00} \\ \hline  
\multicolumn{ 1}{c|}{$f_{11}$}  & 1.015e+04 $\pm$ 4.23e+02 & 5.815e+03 $\pm$ 8.42e+02 & 7.423e+03 $\pm$ 3.28e+02 & \textbf{2.813e+02} $\pm$ \textbf{2.23e+02} \\ \hline  
\multicolumn{ 1}{c|}{$f_{12}$}  & 1.294e+00 $\pm$ 6.90e-01 & 6.646e-04 $\pm$ 1.81e-03 & 3.593e-11 $\pm$ 3.60e-11 & \textbf{0.000e+00} $\pm$ \textbf{0.00e+00} \\ \hline  
\multicolumn{ 1}{c|}{$f_{13}$}  & -9.475e+01 $\pm$ 2.21e+01 & -1.000e+02 $\pm$ 1.20e-03 & -1.000e+02 $\pm$ 1.54e-05 & \textbf{-1.000e+02} $\pm$ \textbf{1.17e-12} \\ \hline  
\multicolumn{ 1}{c|}{$f_{14}$}  & 5.928e-01 $\pm$ 1.34e+00 & 4.295e+00 $\pm$ 5.06e+00 & 5.258e-22 $\pm$ 1.17e-21 & \textbf{4.712e-32} $\pm$ \textbf{5.59e-48} \\ \hline  
\multicolumn{ 1}{c|}{$f_{15}$}  & -1.290e-01 $\pm$ 1.90e+00 & 1.534e+00 $\pm$ 4.07e+00 & \textbf{-1.150e+00} $\pm$ \textbf{6.80e-16} & -1.123e+00 $\pm$ 5.68e-02 \\ \hline  
\multicolumn{ 1}{c|}{$f_{16}$}  & 1.038e+04 $\pm$ 3.36e+03 & 4.376e+03 $\pm$ 1.14e+03 & \textbf{4.291e+03} $\pm$ \textbf{1.01e+03} & 9.660e+03 $\pm$ 2.83e+03 \\ \hline  
\multicolumn{ 1}{c|}{$f_{17}$}  & 4.026e+01 $\pm$ 7.51e-01 & 3.227e+01 $\pm$ 4.11e+00 & 4.011e+01 $\pm$ 8.72e-01 & \textbf{2.762e+01} $\pm$ \textbf{4.43e+00} \\ \hline  
\multicolumn{ 1}{c|}{$f_{18}$}  & 5.277e+05 $\pm$ 2.93e+05 & 4.121e+03 $\pm$ 6.25e+03 & 1.050e+06 $\pm$ 9.99e+04 & \textbf{2.198e+03} $\pm$ \textbf{4.10e+03} \\ \hline  
\multicolumn{ 1}{c|}{$f_{19}$}  & 3.596e+02 $\pm$ 1.37e+01 & \textbf{3.187e+02} $\pm$ \textbf{5.34e+01} & 4.799e+02 $\pm$ 2.42e+01 & 3.853e+02 $\pm$ 4.39e+01 \\ \hline  
\multicolumn{ 1}{c|}{$f_{20}$}  & -1.123e+01 $\pm$ 1.53e+00 & -1.850e+01 $\pm$ 3.48e+00 & -1.068e+01 $\pm$ 3.74e-01 & \textbf{-4.346e+01} $\pm$ \textbf{1.24e+00} \\ \hline  
\multicolumn{ 1}{c|}{$f_{21}$}  & 1.782e+04 $\pm$ 6.93e+02 & 9.592e+03 $\pm$ 1.11e+03 & 1.414e+04 $\pm$ 3.18e+02 & \textbf{1.271e+03} $\pm$ \textbf{4.39e+02} \\ \hline  
\multicolumn{ 1}{c|}{$f_{22}$}  & -1.331e+01 $\pm$ 8.55e-01 & -2.486e+01 $\pm$ 3.01e+00 & -1.354e+01 $\pm$ 7.48e-01 & \textbf{-8.155e+01} $\pm$ \textbf{2.39e+00} \\ \hline  
\multicolumn{ 1}{c|}{$f_{23}$}  & 3.745e+04 $\pm$ 9.82e+02 & 1.990e+04 $\pm$ 1.24e+03 & 3.200e+04 $\pm$ 4.41e+02 & \textbf{3.312e+03} $\pm$ \textbf{4.88e+02} \\ \hline  
\multicolumn{ 1}{c|}{$f_{24}$}  & 3.681e+04 $\pm$ 1.89e+04 & \textbf{1.824e-13} $\pm$ \textbf{2.36e-14} & 3.052e-01 $\pm$ 4.56e-01 & 9.900e-13 $\pm$ 1.70e-13 \\ \hline  
\multicolumn{ 1}{c|}{$f_{25}$}  & 1.220e+01 $\pm$ 2.19e+01 & 6.610e+01 $\pm$ 6.53e+00 & 9.070e+01 $\pm$ 1.37e+01 & \textbf{2.879e-09} $\pm$ \textbf{1.11e-08} \\ \hline  
\multicolumn{ 1}{c|}{$f_{26}$}  & 1.283e+09 $\pm$ 1.28e+09 & \textbf{8.043e+01} $\pm$ \textbf{8.86e+01} & 7.312e+02 $\pm$ 1.49e+03 & 1.432e+02 $\pm$ 1.70e+02 \\ \hline  
\multicolumn{ 1}{c|}{$f_{27}$}  & 9.589e+02 $\pm$ 3.94e+01 & 9.675e+02 $\pm$ 1.20e+02 & 1.352e+03 $\pm$ 8.44e+01 & \textbf{1.132e-12} $\pm$ \textbf{1.86e-13} \\ \hline  
\multicolumn{ 1}{c|}{$f_{28}$}  & 1.904e+03 $\pm$ 1.16e+02 & \textbf{1.745e-03} $\pm$ \textbf{4.33e-03} & 4.197e-02 $\pm$ 6.82e-02 & 2.978e-03 $\pm$ 4.71e-03 \\ \hline  
\multicolumn{ 1}{c|}{$f_{29}$}  & 1.740e+01 $\pm$ 6.12e+00 & 1.988e+01 $\pm$ 3.97e-02 & 8.497e+00 $\pm$ 6.59e+00 & \textbf{2.141e-12} $\pm$ \textbf{2.44e-13} \\ \hline  
\multicolumn{ 1}{c|}{$f_{30}$}  & -8.586e+02 $\pm$ 1.58e+01 & -1.156e+03 $\pm$ 5.07e+01 & -8.541e+02 $\pm$ 8.80e+00 & \textbf{-1.488e+03} $\pm$ \textbf{1.83e+01} \\ \hline  \hline  
\end{tabular} 
\end{tiny}
\end{center} 
\end{table}

\begin{table}[t]
\caption{Wilcoxon test for baseline algorithms}\label{Wilcoxoncomp1} 
\begin{center} 
\begin{scriptsize}
\begin{tabular}{c|c|c|c} 
\hline 
\hline 
\multicolumn{1}{l|}{Test Problem} & EDA$_{mvg}$ & (1+1)-CMA-ES & 2OptDE \\ \hline 
\multicolumn{ 1}{c|}{$f_{1}$}  & + & + & + \\ \hline 
\multicolumn{ 1}{c|}{$f_{2}$}  & + & - & + \\ \hline 
\multicolumn{ 1}{c|}{$f_{3}$}  & + & - & - \\ \hline 
\multicolumn{ 1}{c|}{$f_{4}$}  & + & + & + \\ \hline 
\multicolumn{ 1}{c|}{$f_{5}$}  & - & = & - \\ \hline 
\multicolumn{ 1}{c|}{$f_{6}$}  & + & = & + \\ \hline 
\multicolumn{ 1}{c|}{$f_{7}$}  & + & = & + \\ \hline 
\multicolumn{ 1}{c|}{$f_{8}$}  & + & + & + \\ \hline 
\multicolumn{ 1}{c|}{$f_{9}$}  & - & - & = \\ \hline 
\multicolumn{ 1}{c|}{$f_{10}$}  & + & + & + \\ \hline 
\multicolumn{ 1}{c|}{$f_{11}$}  & + & + & + \\ \hline 
\multicolumn{ 1}{c|}{$f_{12}$}  & + & + & + \\ \hline 
\multicolumn{ 1}{c|}{$f_{13}$}  & + & + & + \\ \hline 
\multicolumn{ 1}{c|}{$f_{14}$}  & + & + & + \\ \hline 
\multicolumn{ 1}{c|}{$f_{15}$}  & + & + & - \\ \hline 
\multicolumn{ 1}{c|}{$f_{16}$}  & = & - & - \\ \hline 
\multicolumn{ 1}{c|}{$f_{17}$}  & + & + & + \\ \hline 
\multicolumn{ 1}{c|}{$f_{18}$}  & + & = & + \\ \hline 
\multicolumn{ 1}{c|}{$f_{19}$}  & - & - & + \\ \hline 
\multicolumn{ 1}{c|}{$f_{20}$}  & + & + & + \\ \hline 
\multicolumn{ 1}{c|}{$f_{21}$}  & + & + & + \\ \hline 
\multicolumn{ 1}{c|}{$f_{22}$}  & + & + & + \\ \hline 
\multicolumn{ 1}{c|}{$f_{23}$}  & + & + & + \\ \hline 
\multicolumn{ 1}{c|}{$f_{24}$}  & + & - & + \\ \hline 
\multicolumn{ 1}{c|}{$f_{25}$}  & + & + & + \\ \hline 
\multicolumn{ 1}{c|}{$f_{26}$}  & + & = & + \\ \hline 
\multicolumn{ 1}{c|}{$f_{27}$}  & + & + & + \\ \hline 
\multicolumn{ 1}{c|}{$f_{28}$}  & + & - & + \\ \hline 
\multicolumn{ 1}{c|}{$f_{29}$}  & + & + & + \\ \hline 
\multicolumn{ 1}{c|}{$f_{30}$}  & + & + & + \\ \hline \hline
\end{tabular} 
\end{scriptsize}
\end{center} 
\end{table} 

Numerical comparisons with these baseline algorithms, displayed in Tables \ref{rescomp1} and \ref{Wilcoxoncomp1}, show that the proposed 3SOME algorithm succeeds at achieving the best performance, among the four algorithms considered, in twenty cases out of the thirty problems under consideration. In addition, the Wilcoxon test shows that 3SOME wins most of the pair-wise comparisons. The most performing competitor, i.e. (1+1)-CMA-ES,  succeeds at outperforming the 3SOME algorithm in only seven cases while it is outperformed in eighteen cases. We can conclude that, for the problems under investigation, the 3SOME algorithm displays the best performance with respect to the other baseline algorithms. 

\begin{figure}[p]
\centering
\subfigure[$f_{11}$\label{fig:11}]{\includegraphics[width=.8\linewidth]{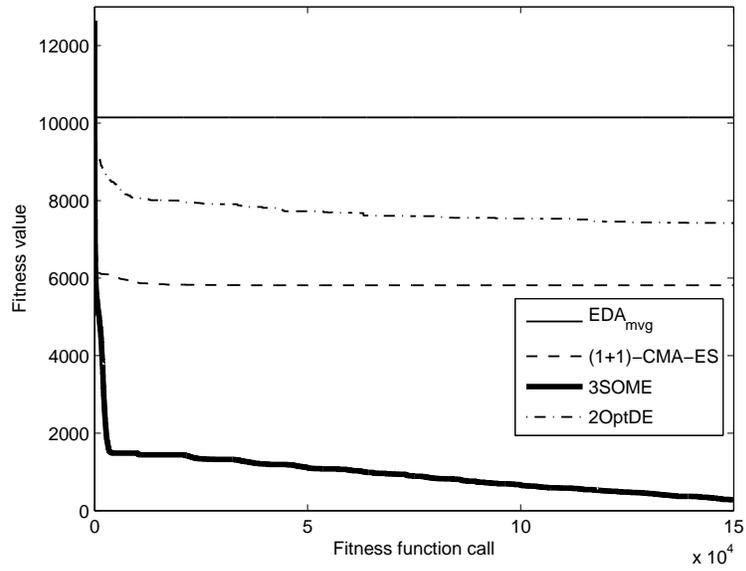}}
\subfigure[$f_{22}$\label{fig:22}]{\includegraphics[width=.8\linewidth]{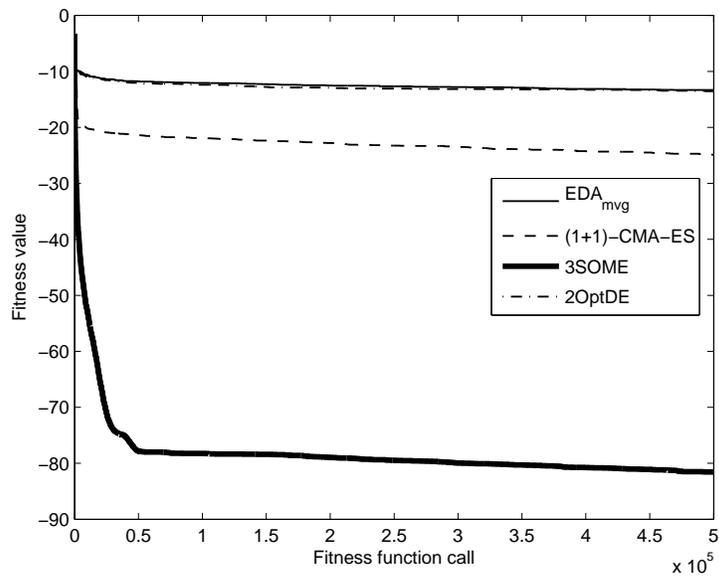}}
\caption{Performance trends of 3SOME and baseline algorithms}\label{fig:perf1}
\end{figure}

Fig. \ref{fig:perf1} shows average performance trends of the four algorithms over a selection of the test problems considered in this study.

It can be observed in Fig. \ref{fig:perf1} that 3SOME is more efficient that the other algorithms at detecting promising basins of attraction since it succeeds very quickly to find high quality solutions, as shown in the steep  trend at the beginning of the optimization. After that phase 3SOME does not stop improving upon the detected solution but continues the optimization towards the global optimum.   

\subsection{Comparison with Memetic Computing Approaches}
The second category is represented by the following algorithms:
\begin{itemize}
\item Real Coded Memetic Algorithm (RCMA) proposed in \cite{bib:Molina2005}. RCMA has been run with population size $N_p=50$, crossover parameter $\alpha=0.5$, mutation probability $0.125$, maximum generation number $T=10000$, $\beta=5$, maximum number of individuals taking part to the negative assortative mating $N_{nam}=3$, roulette wheel selection. The other fixed parameters have been set as suggested in
  \cite{bib:Molina2005}.
  \item  Disturbed Exploitation compact Differential Evolution (DEcDE) proposed in \cite{bib:NeriDEcDE2011}. DEcDE has been run with virtual population size $N_p=300$, scale factor $F=0.5$, inheritance factor $\alpha_m=0.25$, trigonometric mutation probability $M_t=0.003$, and perturbation probability $M_p=0.001$, as suggested in the original paper.
  \item Differential Evolution with adaptive hill-climb Simplex Crossover (DEahcSPX) proposed in \cite{bib:Noman2008}. DEahcSPX has been run with $N_p=50$, scale factor $F=0.9$ and crossover rate $Cr=0.9$ as suggested in the paper, the number of points involved in the hill-climb $n_p=3$, factor  $\epsilon=1$.
\end{itemize} 

Numerical results are given in Tables and \ref{rescomp2} and \ref{Wilcoxoncomp2} .

\begin{table}[t]
\caption{Fitness comparison with memetic computing approaches} \label{rescomp2}
\begin{center} 
\begin{tiny}
\begin{tabular}{c|c|c|c|c} 
\hline 
\hline  
\multicolumn{1}{l|}{Test Problem} & RCMA & DEcDE & DEahcSPX & 3SOME \\ \hline  
\multicolumn{ 1}{c|}{$f_{1}$}  & 3.411e-16 $\pm$ 6.85e-16 & 1.530e+00 $\pm$ 2.58e+00 & 3.907e-20 $\pm$ 1.81e-20 & \textbf{0.000e+00} $\pm$ \textbf{0.00e+00} \\ \hline  
\multicolumn{ 1}{c|}{$f_{2}$}  & 1.255e-13 $\pm$ 2.95e-13 & 1.037e+03 $\pm$ 1.09e+03 & 8.878e+01 $\pm$ 3.86e+01 & \textbf{1.604e-23} $\pm$ \textbf{2.99e-23} \\ \hline  
\multicolumn{ 1}{c|}{$f_{3}$}  & \textbf{2.827e+01} $\pm$ \textbf{1.28e-01} & 8.628e+02 $\pm$ 6.70e+02 & 4.310e+01 $\pm$ 1.95e+01 & 5.393e+01 $\pm$ 1.60e+02 \\ \hline  
\multicolumn{ 1}{c|}{$f_{4}$}  & 7.012e-09 $\pm$ 1.04e-08 & 6.058e-01 $\pm$ 4.78e-01 & 5.415e-11 $\pm$ 1.20e-11 & \textbf{4.796e-14} $\pm$ \textbf{1.21e-14} \\ \hline  
\multicolumn{ 1}{c|}{$f_{5}$}  & \textbf{8.119e-09} $\pm$ \textbf{8.45e-09} & 2.482e+00 $\pm$ 5.10e-01 & 1.460e-01 $\pm$ 3.63e-02 & 1.645e+01 $\pm$ 6.39e+00 \\ \hline  
\multicolumn{ 1}{c|}{$f_{6}$}  & 6.575e+00 $\pm$ 3.85e+00 & 2.763e-01 $\pm$ 2.36e-01 & \textbf{6.166e-09} $\pm$ \textbf{1.13e-08} & 4.313e-03 $\pm$ 4.55e-03 \\ \hline  
\multicolumn{ 1}{c|}{$f_{7}$}  & 6.063e+00 $\pm$ 3.50e+00 & 3.957e-01 $\pm$ 2.31e-01 & \textbf{3.126e-02} $\pm$ \textbf{5.71e-02} & 1.788e-01 $\pm$ 1.89e-01 \\ \hline  
\multicolumn{ 1}{c|}{$f_{8}$}  & 7.105e-15 $\pm$ 2.55e-14 & 1.796e+01 $\pm$ 4.01e+00 & \textbf{0.000e+00} $\pm$ \textbf{0.00e+00} & 2.487e-13 $\pm$ 1.16e-13 \\ \hline  
\multicolumn{ 1}{c|}{$f_{9}$}  & \textbf{4.737e-15} $\pm$ \textbf{2.32e-14} & 2.065e+02 $\pm$ 2.99e+01 & 1.115e+02 $\pm$ 1.92e+01 & 2.281e+02 $\pm$ 4.76e+01 \\ \hline  
\multicolumn{ 1}{c|}{$f_{10}$}  & 1.003e+04 $\pm$ 1.78e+04 & 1.997e+03 $\pm$ 4.00e+02 & \textbf{1.073e-12} $\pm$ \textbf{8.01e-13} & 3.661e+01 $\pm$ 8.03e+00 \\ \hline  
\multicolumn{ 1}{c|}{$f_{11}$}  & 2.989e+03 $\pm$ 5.98e+02 & 4.163e+02 $\pm$ 1.19e+02 & \textbf{3.818e-04} $\pm$ \textbf{3.71e-13} & 2.813e+02 $\pm$ 2.23e+02 \\ \hline  
\multicolumn{ 1}{c|}{$f_{12}$}  & 6.329e-10 $\pm$ 2.63e-09 & 5.229e-02 $\pm$ 3.76e-02 & 1.668e-09 $\pm$ 5.92e-10 & \textbf{0.000e+00} $\pm$ \textbf{0.00e+00} \\ \hline  
\multicolumn{ 1}{c|}{$f_{13}$}  & -6.845e+01 $\pm$ 1.17e+01 & -9.975e+01 $\pm$ 1.53e-01 & -9.882e+01 $\pm$ 1.30e-01 & \textbf{-1.000e+02} $\pm$ \textbf{1.17e-12} \\ \hline  
\multicolumn{ 1}{c|}{$f_{14}$}  & 1.030e-06 $\pm$ 1.29e-06 & 1.565e-03 $\pm$ 2.13e-03 & 2.021e-17 $\pm$ 1.77e-17 & \textbf{4.712e-32} $\pm$ \textbf{5.59e-48} \\ \hline  
\multicolumn{ 1}{c|}{$f_{15}$}  & -1.149e+00 $\pm$ 3.70e-03 & -1.119e+00 $\pm$ 4.18e-02 & \textbf{-1.150e+00} $\pm$ \textbf{4.09e-16} & -1.123e+00 $\pm$ 5.68e-02 \\ \hline  
\multicolumn{ 1}{c|}{$f_{16}$}  & 9.628e+03 $\pm$ 2.45e+03 & 5.945e+03 $\pm$ 1.66e+03 & \textbf{5.649e+03} $\pm$ \textbf{4.03e+02} & 9.660e+03 $\pm$ 2.83e+03 \\ \hline  
\multicolumn{ 1}{c|}{$f_{17}$}  & 3.556e+01 $\pm$ 3.65e+00 & 3.941e+01 $\pm$ 1.40e+00 & 3.973e+01 $\pm$ 1.44e+00 & \textbf{2.762e+01} $\pm$ \textbf{4.43e+00} \\ \hline  
\multicolumn{ 1}{c|}{$f_{18}$}  & 1.302e+05 $\pm$ 4.67e+04 & 1.282e+05 $\pm$ 2.53e+04 & 7.186e+04 $\pm$ 1.49e+04 & \textbf{2.198e+03} $\pm$ \textbf{4.10e+03} \\ \hline  
\multicolumn{ 1}{c|}{$f_{19}$}  & \textbf{4.577e-10} $\pm$ \textbf{4.46e-10} & 4.343e+02 $\pm$ 2.18e+01 & 2.698e+02 $\pm$ 2.99e+01 & 3.853e+02 $\pm$ 4.39e+01 \\ \hline  
\multicolumn{ 1}{c|}{$f_{20}$}  & -2.641e+01 $\pm$ 2.23e+00 & -3.071e+01 $\pm$ 2.46e+00 & -4.237e+01 $\pm$ 4.61e-01 & \textbf{-4.346e+01} $\pm$ \textbf{1.24e+00} \\ \hline  
\multicolumn{ 1}{c|}{$f_{21}$}  & 7.706e+03 $\pm$ 7.52e+02 & 2.391e+03 $\pm$ 3.25e+02 & \textbf{6.364e-04} $\pm$ \textbf{0.00e+00} & 1.271e+03 $\pm$ 4.39e+02 \\ \hline  
\multicolumn{ 1}{c|}{$f_{22}$}  & -3.116e+01 $\pm$ 2.59e+00 & -3.838e+01 $\pm$ 2.44e+00 & -5.364e+01 $\pm$ 9.60e-01 & \textbf{-8.155e+01} $\pm$ \textbf{2.39e+00} \\ \hline  
\multicolumn{ 1}{c|}{$f_{23}$}  & 2.055e+04 $\pm$ 1.27e+03 & 1.113e+04 $\pm$ 8.62e+02 & \textbf{1.514e-03} $\pm$ \textbf{1.19e-04} & 3.312e+03 $\pm$ 4.88e+02 \\ \hline  
\multicolumn{ 1}{c|}{$f_{24}$}  & 2.571e+04 $\pm$ 5.01e+03 & 2.408e+03 $\pm$ 8.00e+02 & 8.562e-09 $\pm$ 2.24e-09 & \textbf{9.900e-13} $\pm$ \textbf{1.70e-13} \\ \hline  
\multicolumn{ 1}{c|}{$f_{25}$}  & 6.194e-01 $\pm$ 2.66e+00 & 6.887e+01 $\pm$ 3.60e+00 & 2.388e+01 $\pm$ 1.21e+00 & \textbf{2.879e-09} $\pm$ \textbf{1.11e-08} \\ \hline  
\multicolumn{ 1}{c|}{$f_{26}$}  & 1.873e+09 $\pm$ 7.29e+08 & 3.046e+07 $\pm$ 1.56e+07 & 3.854e+02 $\pm$ 6.26e+01 & \textbf{1.432e+02} $\pm$ \textbf{1.70e+02} \\ \hline  
\multicolumn{ 1}{c|}{$f_{27}$}  & 7.820e+02 $\pm$ 9.68e+01 & 4.179e+02 $\pm$ 2.10e+01 & 1.870e+02 $\pm$ 1.23e+01 & \textbf{1.132e-12} $\pm$ \textbf{1.86e-13} \\ \hline  
\multicolumn{ 1}{c|}{$f_{28}$}  & 1.902e+02 $\pm$ 3.97e+01 & 2.542e+01 $\pm$ 7.34e+00 & \textbf{4.917e-09} $\pm$ \textbf{1.77e-09} & 2.978e-03 $\pm$ 4.71e-03 \\ \hline  
\multicolumn{ 1}{c|}{$f_{29}$}  & 1.934e+01 $\pm$ 8.46e-01 & 7.488e+00 $\pm$ 8.78e-01 & 1.081e-05 $\pm$ 1.51e-06 & \textbf{2.141e-12} $\pm$ \textbf{2.44e-13} \\ \hline  
\multicolumn{ 1}{c|}{$f_{30}$}  & -1.120e+03 $\pm$ 3.80e+01 & -1.286e+03 $\pm$ 1.80e+01 & -1.197e+03 $\pm$ 1.02e+01 & \textbf{-1.488e+03} $\pm$ \textbf{1.83e+01} \\ \hline  \hline  
\end{tabular} 
\end{tiny}
\end{center} 
\end{table}

\begin{table}[t]
\caption{Wilcoxon test for memetic computing approaches}\label{Wilcoxoncomp2} 
\begin{center} 
\begin{scriptsize}
\begin{tabular}{c|c|c|c} 
\hline 
\hline 
\multicolumn{1}{l|}{Test Problem} & RCMA & DEcDE & DEahcSPX \\ \hline 
\multicolumn{ 1}{c|}{$f_{1}$}  & + & + & + \\ \hline 
\multicolumn{ 1}{c|}{$f_{2}$}  & + & + & + \\ \hline 
\multicolumn{ 1}{c|}{$f_{3}$}  & - & + & - \\ \hline 
\multicolumn{ 1}{c|}{$f_{4}$}  & + & + & + \\ \hline 
\multicolumn{ 1}{c|}{$f_{5}$}  & - & - & - \\ \hline 
\multicolumn{ 1}{c|}{$f_{6}$}  & + & + & = \\ \hline 
\multicolumn{ 1}{c|}{$f_{7}$}  & + & + & = \\ \hline 
\multicolumn{ 1}{c|}{$f_{8}$}  & - & + & - \\ \hline 
\multicolumn{ 1}{c|}{$f_{9}$}  & - & - & - \\ \hline 
\multicolumn{ 1}{c|}{$f_{10}$}  & + & + & - \\ \hline 
\multicolumn{ 1}{c|}{$f_{11}$}  & + & + & - \\ \hline 
\multicolumn{ 1}{c|}{$f_{12}$}  & + & + & + \\ \hline 
\multicolumn{ 1}{c|}{$f_{13}$}  & + & + & + \\ \hline 
\multicolumn{ 1}{c|}{$f_{14}$}  & + & + & + \\ \hline 
\multicolumn{ 1}{c|}{$f_{15}$}  & - & + & = \\ \hline 
\multicolumn{ 1}{c|}{$f_{16}$}  & = & - & - \\ \hline 
\multicolumn{ 1}{c|}{$f_{17}$}  & + & + & + \\ \hline 
\multicolumn{ 1}{c|}{$f_{18}$}  & + & + & + \\ \hline 
\multicolumn{ 1}{c|}{$f_{19}$}  & - & + & - \\ \hline 
\multicolumn{ 1}{c|}{$f_{20}$}  & + & + & + \\ \hline 
\multicolumn{ 1}{c|}{$f_{21}$}  & + & + & - \\ \hline 
\multicolumn{ 1}{c|}{$f_{22}$}  & + & + & + \\ \hline 
\multicolumn{ 1}{c|}{$f_{23}$}  & + & + & - \\ \hline 
\multicolumn{ 1}{c|}{$f_{24}$}  & + & + & + \\ \hline 
\multicolumn{ 1}{c|}{$f_{25}$}  & + & + & + \\ \hline 
\multicolumn{ 1}{c|}{$f_{26}$}  & + & + & + \\ \hline 
\multicolumn{ 1}{c|}{$f_{27}$}  & + & + & + \\ \hline 
\multicolumn{ 1}{c|}{$f_{28}$}  & + & + & - \\ \hline 
\multicolumn{ 1}{c|}{$f_{29}$}  & + & + & + \\ \hline 
\multicolumn{ 1}{c|}{$f_{30}$}  & + & + & + \\ \hline \hline
\end{tabular} 
\end{scriptsize}
\end{center} 
\end{table}

Numerical results in Table \ref{rescomp2} and \ref{Wilcoxoncomp2} show that the proposed 3SOME displays a competitive performance with modern MA. It can easlily be observed that 3SOME outperforms all the MC approaches in sixteen cases out of the thirty problems under consideration.  DEahcSPX also displays  a respectable performance. However, the pairwise comparison obtained by Wilcoxon test shows that DEahcSPX outperforms 3SOME in eleven cases and is outperformed by 3SOME in sixteen cases. Thus, 3SOME appears to have the best performance with respect to the three MAs considered in this subsection.   

\begin{figure}[p]
\centering
\subfigure[$f_{17}$\label{fig:17}]{\includegraphics[width=.8\linewidth]{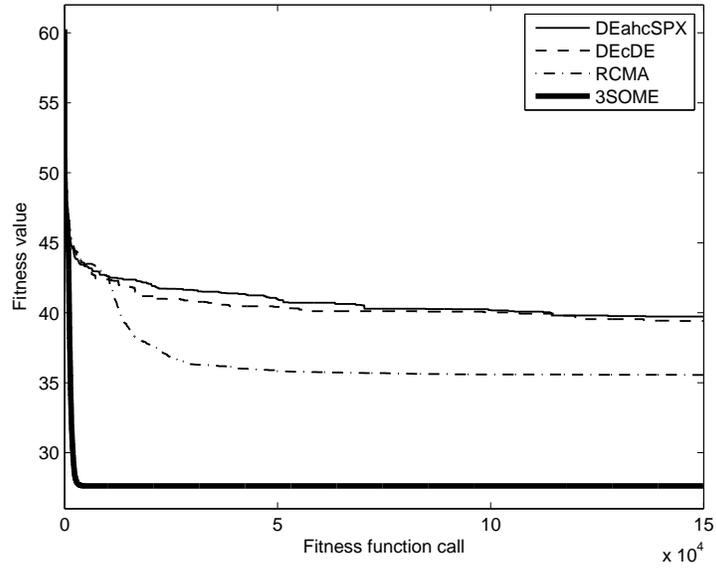}}
\subfigure[$f_{27}$\label{fig:27}]{\includegraphics[width=.8\linewidth]{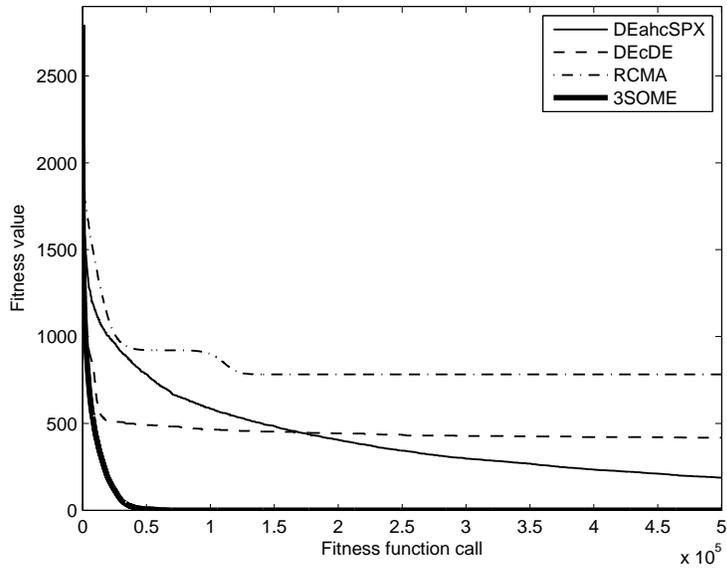}}
\caption{Performance trends of 3SOME and memetic computing approaches}\label{fig:perf2}
\end{figure}

Fig. \ref{fig:perf2} shows average performance trends of the four algorithms  over a selection of the test problems considered in this study.

Graphical results in Fig. \ref{fig:perf2} show that the 3SOME algorithm for some landscapes is very efficient at generating high performance solutions at the very beginning of the optimization process.

\subsection{Comparison with complex optimization algorithms}
The third category is represented by the following algorithms:
\begin{itemize}
  \item Self-Adaptive Differential Evolution (SADE) proposed in \cite{bib:Qin2009}. SADE has been run with Learning Period $LP=20$ and population size $N_p=50$. The other constant values are the same reported in the formulas of the original paper.  
  \item Differential Evolution with Global and Local -based mutation (DEGL) proposed in \cite{bib:Das2009}. DEGL has been run  with population size $N_p=50$, scale factor $F=0.75$ and crossover rate $Cr=0.3$, and parameter $k=10$, as suggested in the original paper. 
  \item Frankenstein PSO proposed in \cite{bib:deOca2009}. The topology update parameter is $k=2000$ iterations, inertia weight schedule $\omega t_{max}$, $\omega_{max}=0.969$,  $\omega_{min}=0.175$, and maximum velocity $v_{max}=0.75$, i.e. velocity can vary between $-0.75$ and $0.75$. The population size (swarm size in this case) is $N_p=50$ as for the other population based algorithms.  
\end{itemize} 

Numerical results are given in Tables and \ref{rescomp3} and \ref{Wilcoxoncomp3} .

\begin{table}[t]
\caption{Fitness comparison with the-state-of-the-art complex algorithms} \label{rescomp3}
\begin{center} 
\begin{tiny}
\begin{tabular}{c|c|c|c|c} 
\hline 
\hline  
\multicolumn{1}{l|}{Test Problem} & SADE & DEGL & FrankensteinPSO & 3SOME \\ \hline  
\multicolumn{ 1}{c|}{$f_{1}$}  & 2.137e-27 $\pm$ 3.61e-27 & 6.502e-76 $\pm$ 1.76e-75 & 3.388e-11 $\pm$ 1.11e-10 & \textbf{0.000e+00} $\pm$ \textbf{0.00e+00} \\ \hline  
\multicolumn{ 1}{c|}{$f_{2}$}  & 1.768e-02 $\pm$ 3.58e-02 & 4.020e-03 $\pm$ 5.12e-03 & 2.509e+02 $\pm$ 2.88e+02 & \textbf{1.604e-23} $\pm$ \textbf{2.99e-23} \\ \hline  
\multicolumn{ 1}{c|}{$f_{3}$}  & 1.148e+01 $\pm$ 3.88e+00 & \textbf{5.389e-01} $\pm$ \textbf{1.47e+00} & 1.306e+02 $\pm$ 3.24e+02 & 5.393e+01 $\pm$ 1.60e+02 \\ \hline  
\multicolumn{ 1}{c|}{$f_{4}$}  & \textbf{6.661e-15} $\pm$ \textbf{2.53e-15} & 1.803e-01 $\pm$ 4.90e-01 & 1.655e-07 $\pm$ 4.49e-07 & 4.796e-14 $\pm$ 1.21e-14 \\ \hline  
\multicolumn{ 1}{c|}{$f_{5}$}  & \textbf{8.290e-15} $\pm$ \textbf{3.13e-15} & 7.761e-02 $\pm$ 2.63e-01 & 1.680e-06 $\pm$ 7.07e-06 & 1.645e+01 $\pm$ 6.39e+00 \\ \hline  
\multicolumn{ 1}{c|}{$f_{6}$}  & \textbf{0.000e+00} $\pm$ \textbf{0.00e+00} & 9.954e-02 $\pm$ 1.72e-01 & 1.425e+02 $\pm$ 1.36e+01 & 4.313e-03 $\pm$ 4.55e-03 \\ \hline  
\multicolumn{ 1}{c|}{$f_{7}$}  & \textbf{2.676e-02} $\pm$ \textbf{7.59e-02} & 1.278e-01 $\pm$ 1.91e-01 & 1.452e+02 $\pm$ 1.04e+01 & 1.788e-01 $\pm$ 1.89e-01 \\ \hline  
\multicolumn{ 1}{c|}{$f_{8}$}  & 1.965e+01 $\pm$ 1.51e+01 & 8.211e+01 $\pm$ 5.37e+01 & 6.882e+00 $\pm$ 2.95e+00 & \textbf{2.487e-13} $\pm$ \textbf{1.16e-13} \\ \hline  
\multicolumn{ 1}{c|}{$f_{9}$}  & 2.960e+01 $\pm$ 1.08e+01 & 1.763e+02 $\pm$ 1.68e+01 & \textbf{7.504e+00} $\pm$ \textbf{2.78e+00} & 2.281e+02 $\pm$ 4.76e+01 \\ \hline  
\multicolumn{ 1}{c|}{$f_{10}$}  & \textbf{1.749e+01} $\pm$ \textbf{1.43e+01} & 1.013e+02 $\pm$ 4.89e+01 & 7.745e+05 $\pm$ 8.84e+04 & 3.661e+01 $\pm$ 8.03e+00 \\ \hline  
\multicolumn{ 1}{c|}{$f_{11}$}  & 1.391e+03 $\pm$ 1.14e+03 & 2.112e+03 $\pm$ 1.24e+03 & 8.856e+03 $\pm$ 3.73e+02 & \textbf{2.813e+02} $\pm$ \textbf{2.23e+02} \\ \hline  
\multicolumn{ 1}{c|}{$f_{12}$}  & \textbf{0.000e+00} $\pm$ \textbf{0.00e+00} & 4.403e-32 $\pm$ 6.37e-32 & 1.431e-12 $\pm$ 5.54e-13 & \textbf{0.000e+00} $\pm$ \textbf{0.00e+00} \\ \hline  
\multicolumn{ 1}{c|}{$f_{13}$}  & \textbf{-1.000e+02} $\pm$ \textbf{0.00e+00} & \textbf{-1.000e+02} $\pm$ \textbf{0.00e+00} & -5.496e+01 $\pm$ 4.93e+00 & -1.000e+02 $\pm$ 1.17e-12 \\ \hline  
\multicolumn{ 1}{c|}{$f_{14}$}  & \textbf{4.712e-32} $\pm$ \textbf{5.59e-48} & \textbf{4.712e-32} $\pm$ \textbf{5.59e-48} & 1.060e-24 $\pm$ 2.04e-24 & \textbf{4.712e-32} $\pm$ \textbf{5.59e-48} \\ \hline  
\multicolumn{ 1}{c|}{$f_{15}$}  & \textbf{-1.150e+00} $\pm$ \textbf{6.80e-16} & \textbf{-1.150e+00} $\pm$ \textbf{6.80e-16} & \textbf{-1.150e+00} $\pm$ \textbf{6.80e-16} & -1.123e+00 $\pm$ 5.68e-02 \\ \hline  
\multicolumn{ 1}{c|}{$f_{16}$}  & 2.174e+03 $\pm$ 5.61e+02 & \textbf{3.626e+02} $\pm$ \textbf{6.82e+02} & 1.430e+04 $\pm$ 8.48e+02 & 9.660e+03 $\pm$ 2.83e+03 \\ \hline  
\multicolumn{ 1}{c|}{$f_{17}$}  & 3.463e+01 $\pm$ 6.57e+00 & 3.988e+01 $\pm$ 1.21e+00 & 4.011e+01 $\pm$ 1.58e+00 & \textbf{2.762e+01} $\pm$ \textbf{4.43e+00} \\ \hline  
\multicolumn{ 1}{c|}{$f_{18}$}  & 2.819e+04 $\pm$ 2.45e+04 & 1.002e+06 $\pm$ 1.99e+05 & 1.088e+06 $\pm$ 1.14e+05 & \textbf{2.198e+03} $\pm$ \textbf{4.10e+03} \\ \hline  
\multicolumn{ 1}{c|}{$f_{19}$}  & 4.718e+01 $\pm$ 1.08e+01 & 3.569e+02 $\pm$ 1.87e+01 & \textbf{1.552e+01} $\pm$ \textbf{4.10e+00} & 3.853e+02 $\pm$ 4.39e+01 \\ \hline  
\multicolumn{ 1}{c|}{$f_{20}$}  & -4.241e+01 $\pm$ 2.11e+00 & -1.384e+01 $\pm$ 6.33e-01 & -1.987e+01 $\pm$ 4.59e+00 & \textbf{-4.346e+01} $\pm$ \textbf{1.24e+00} \\ \hline  
\multicolumn{ 1}{c|}{$f_{21}$}  & 3.040e+03 $\pm$ 2.21e+03 & 3.641e+03 $\pm$ 7.54e+02 & 1.639e+04 $\pm$ 4.53e+02 & \textbf{1.271e+03} $\pm$ \textbf{4.39e+02} \\ \hline  
\multicolumn{ 1}{c|}{$f_{22}$}  & -8.013e+01 $\pm$ 3.16e+00 & -1.652e+01 $\pm$ 9.52e-01 & -2.276e+01 $\pm$ 4.45e+00 & \textbf{-8.155e+01} $\pm$ \textbf{2.39e+00} \\ \hline  
\multicolumn{ 1}{c|}{$f_{23}$}  & 1.245e+04 $\pm$ 2.07e+03 & 9.199e+03 $\pm$ 1.68e+03 & 3.518e+04 $\pm$ 6.83e+02 & \textbf{3.312e+03} $\pm$ \textbf{4.88e+02} \\ \hline  
\multicolumn{ 1}{c|}{$f_{24}$}  & 5.803e-13 $\pm$ 5.79e-13 & \textbf{1.326e-13} $\pm$ \textbf{3.99e-14} & 8.711e+04 $\pm$ 8.28e+03 & 9.900e-13 $\pm$ 1.70e-13 \\ \hline  
\multicolumn{ 1}{c|}{$f_{25}$}  & 7.389e+01 $\pm$ 6.51e+00 & 4.922e+01 $\pm$ 5.26e+00 & 8.006e+01 $\pm$ 3.09e+00 & \textbf{2.879e-09} $\pm$ \textbf{1.11e-08} \\ \hline  
\multicolumn{ 1}{c|}{$f_{26}$}  & 1.713e+02 $\pm$ 9.20e+01 & \textbf{8.094e+01} $\pm$ \textbf{5.36e+01} & 1.042e+10 $\pm$ 1.93e+09 & 1.432e+02 $\pm$ 1.70e+02 \\ \hline  
\multicolumn{ 1}{c|}{$f_{27}$}  & 1.737e+02 $\pm$ 5.35e+01 & 2.132e+02 $\pm$ 6.15e+01 & 9.526e+02 $\pm$ 6.34e+01 & \textbf{1.132e-12} $\pm$ \textbf{1.86e-13} \\ \hline  
\multicolumn{ 1}{c|}{$f_{28}$}  & 6.865e-03 $\pm$ 1.10e-02 & 1.835e-01 $\pm$ 7.10e-01 & 6.242e+02 $\pm$ 4.93e+01 & \textbf{2.978e-03} $\pm$ \textbf{4.71e-03} \\ \hline  
\multicolumn{ 1}{c|}{$f_{29}$}  & 1.595e+00 $\pm$ 8.68e-01 & 3.363e+00 $\pm$ 2.77e+00 & 1.743e+01 $\pm$ 2.66e-01 & \textbf{2.141e-12} $\pm$ \textbf{2.44e-13} \\ \hline  
\multicolumn{ 1}{c|}{$f_{30}$}  & -1.358e+03 $\pm$ 3.52e+01 & -9.239e+02 $\pm$ 4.89e+01 & -1.112e+03 $\pm$ 9.95e+01 & \textbf{-1.488e+03} $\pm$ \textbf{1.83e+01} \\ \hline  \hline  
\end{tabular} 
\end{tiny}
\end{center} 
\end{table} 

\begin{table}[t]
\caption{Wilcoxon test for the-state-of-the-art complex  algorithms}\label{Wilcoxoncomp3} 
\begin{center} 
\begin{scriptsize}
\begin{tabular}{c|c|c|c} 
\hline 
\hline 
\multicolumn{1}{l|}{Test Problem} & SADE & DEGL & FrankensteinPSO \\ \hline 
\multicolumn{ 1}{c|}{$f_{1}$}  & + & + & + \\ \hline 
\multicolumn{ 1}{c|}{$f_{2}$}  & + & + & + \\ \hline 
\multicolumn{ 1}{c|}{$f_{3}$}  & = & - & + \\ \hline 
\multicolumn{ 1}{c|}{$f_{4}$}  & - & + & + \\ \hline 
\multicolumn{ 1}{c|}{$f_{5}$}  & - & - & - \\ \hline 
\multicolumn{ 1}{c|}{$f_{6}$}  & - & = & + \\ \hline 
\multicolumn{ 1}{c|}{$f_{7}$}  & - & - & + \\ \hline 
\multicolumn{ 1}{c|}{$f_{8}$}  & + & + & + \\ \hline 
\multicolumn{ 1}{c|}{$f_{9}$}  & - & - & - \\ \hline 
\multicolumn{ 1}{c|}{$f_{10}$}  & - & + & + \\ \hline 
\multicolumn{ 1}{c|}{$f_{11}$}  & + & + & + \\ \hline 
\multicolumn{ 1}{c|}{$f_{12}$}  & = & + & + \\ \hline 
\multicolumn{ 1}{c|}{$f_{13}$}  & - & - & + \\ \hline 
\multicolumn{ 1}{c|}{$f_{14}$}  & = & = & + \\ \hline 
\multicolumn{ 1}{c|}{$f_{15}$}  & - & - & - \\ \hline 
\multicolumn{ 1}{c|}{$f_{16}$}  & - & - & + \\ \hline 
\multicolumn{ 1}{c|}{$f_{17}$}  & + & + & + \\ \hline 
\multicolumn{ 1}{c|}{$f_{18}$}  & + & + & + \\ \hline 
\multicolumn{ 1}{c|}{$f_{19}$}  & - & - & - \\ \hline 
\multicolumn{ 1}{c|}{$f_{20}$}  & = & + & + \\ \hline 
\multicolumn{ 1}{c|}{$f_{21}$}  & + & + & + \\ \hline 
\multicolumn{ 1}{c|}{$f_{22}$}  & = & + & + \\ \hline 
\multicolumn{ 1}{c|}{$f_{23}$}  & + & + & + \\ \hline 
\multicolumn{ 1}{c|}{$f_{24}$}  & - & - & + \\ \hline 
\multicolumn{ 1}{c|}{$f_{25}$}  & + & + & + \\ \hline 
\multicolumn{ 1}{c|}{$f_{26}$}  & + & = & + \\ \hline 
\multicolumn{ 1}{c|}{$f_{27}$}  & + & + & + \\ \hline 
\multicolumn{ 1}{c|}{$f_{28}$}  & = & = & + \\ \hline 
\multicolumn{ 1}{c|}{$f_{29}$}  & + & + & + \\ \hline 
\multicolumn{ 1}{c|}{$f_{30}$}  & + & + & + \\ \hline \hline
\end{tabular} 
\end{scriptsize}
\end{center} 
\end{table}

Results in Tables \ref{rescomp3} and \ref{Wilcoxoncomp3} prove that the proposed 3SOME algorithm, despite its simplicity succeeds at displaying a respectable performance also against complex algorithms recently proposed in literature. It can be observed from Table \ref{rescomp3}  that the 3SOME algorithm obtains the best performance in seventeen cases out of thirty test problems under consideration. The SADE algorithm appears to be relatively competitive with 3SOME since it obtained the best performance in eleven cases.  In addition, as shown in Table \ref{Wilcoxoncomp3}, 3SOME significantly outperformed SADE in thirteen cases and is outperformed in eleven cases. SADE is a good example of complex algorithm, designed on the basis of DE framework and includes a database for performing a learning of the problem, multiple mutation schemes, and adaptive parameters which are periodically sampled from distribution in order to automatically select the proper mutation scheme and control parameters.  It is important to notice that SADE displays the best performance for $10$ and $30$ dimensional problems, i.e. those problems for which it has been tuned, see \cite{bib:Qin2009}. On the other hand 3SOME seems to offer a better performance than SADE in high dimensional cases. This fact occurs regardless the fact that 3SOME, in our design intentions, does not contain specific components for handling large scale problems. This observation, according to our interpretation, can be read as a confirmation that simple algorithms composed of few and simple memes can be powerful optimization instruments and perform as well as overwhelmingly complex algorithms.   

\begin{figure}[p]
\centering
\subfigure[$f_{25}$\label{fig:25}]{\includegraphics[width=.8\linewidth]{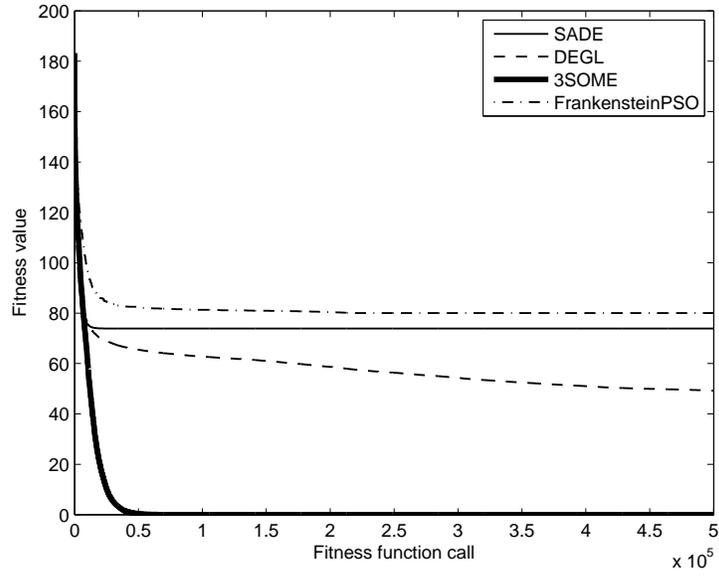}}
\subfigure[$f_{30}$\label{fig:30}]{\includegraphics[width=.8\linewidth]{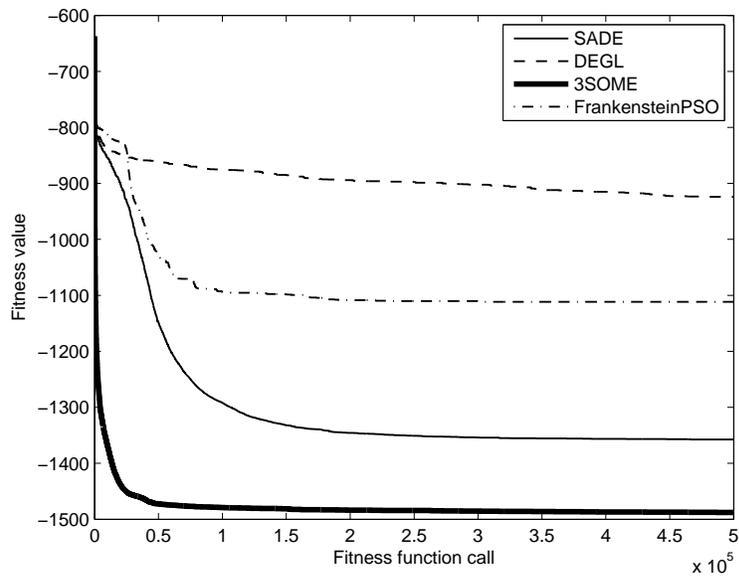}}
\caption{Performance trends of 3SOME and  the-state-of-the-art complex algorithms}\label{fig:perf3}
\end{figure}

Fig. \ref{fig:perf3} shows average performance trends of the four  algorithms over a selection of the test problems considered in this study.

\subsection{Statistical ranking of the set of algorithms}

In order to draw some statistically significant conclusions regarding the performance of the 3SOME algorithm, the Holm procedure, see \cite{bib:Holm1979} and \cite{bib:Garcia2008}, for the ten algorithms and thirty problems under consideration has been performed. The Holm procedure consists of the following. Considering the results in the tables above, the ten algorithms under analysis have been ranked on the basis of their average
performance calculated over the thirty test problems. More specifically, a score $R_i$ for $i = 1,\dots,N_A$ (where $N_A$ is the number of algorithms under analysis, $N_A = 10$ in our case) has been assigned. The score has been assigned in the following way: for each problem, a score of $10$ is assigned to the algorithm displaying the best performance, $9$ is assigned to the second best, $8$ to the third and so on. The algorithm displaying the worst performance scores $1$. For each algorithm, the scores outcoming by each problem are summed up averaged over the amount of test problems ($30$ in our case). On the basis of these scores the algorithms are sorted (ranked).  With the calculated $R_i$ values,  3SOME has been taken as a reference algorithm. Indicating with $R_0$ the rank of 3SOME, and with $R_j$ for $j = 1,\dots,N_A-1$ the rank of one of the remaining nine algorithms, the values $z_j$ have been calculated as \[ z_j = \frac{R_j - R_0}{\sqrt{\frac{N_A(N_A+1)}{6N_{TP}}}} \] where $N_{TP}$ is the number of test problems in consideration ($N_{TP} = 30$ in our case). By means of the $z_j$ values, the corresponding cumulative normal distribution values $p_j$ have been calculated. These $p_j$ values have then been compared with the corresponding $\delta / (N_A - j)$ where $\delta$ is the level of confidence, set to $0.05$ in our case. Table \ref{holm-test} displays  $z_j$ values, $p_j$ values, and corresponding $\delta/(N_A-j)$. The values of $z_j$ and $p_j$ are expressed in terms of $z_{N_A-j}$ and $p_{N_A-j}$ for $j=1,\dots, N_A - 1$. Moreover, it is indicated whether the null-hypothesis (that the two algorithms have indistinguishable performances) is ``Rejected'' i.e.,  3SOME statistically outperforms the
algorithm under consideration, or ``Accepted'' if the distribution of values can be considered the same (there is no outperformance).

\begin{table}[t]
\caption{Holm procedure for the entire testbed and algorithmic set} \label{holm-test}
\begin{center} 
\begin{scriptsize}
\begin{tabular}{ c c c c c c } 
\hline \hline
${N_A-j}$ & Algorithm & ${z_{N_A}-j}$ & ${p_{N_A}-j}$ & $\delta/(N_A-j)$ & Hypothesis \\ \hline \hline 
9 & EDA$_{mvg}$ & -6.95e+00 & 1.82e-12 & 5.56e-03 & Rejected \\ \hline  
8 & FrankensteinPSO & -5.12e+00 & 1.55e-07 & 6.25e-03 & Rejected \\ \hline  
7 & 2OptDE & -4.95e+00 & 3.78e-07 & 7.14e-03 & Rejected \\ \hline  
6 & DEcDE & -3.92e+00 & 4.37e-05 & 8.33e-03 & Rejected \\ \hline  
5 & (1+1)-CMA-ES & -3.11e+00 & 9.27e-04 & 1.00e-02 & Rejected \\ \hline  
4 & RCMA & -3.07e+00 & 1.07e-03 & 1.25e-02 & Rejected \\ \hline  
3 & DEGL & -1.66e+00 & 4.82e-02 & 1.67e-02 & Accepted \\ \hline  
2 & DEahcSPX & -6.40e-01 & 2.61e-01 & 2.50e-02 & Accepted \\ \hline  
1 & SADE & 2.13e-01 & 5.84e-01 & 5.00e-02 & Accepted \\ \hline  \hline
\end{tabular} 
\end{scriptsize}
\end{center} 
\end{table} 

Numerical results show that the proposed 3SOME is a promising algorithm, since it outperforms, on average, six out nine algorithms considered in this study. In addition, it can be observed that 3SOME displays a performance, on average as good as that of DEGL, DEahcSPX, and SADE. The latter algorithms are well-known successful optimization algorithms which have been recently proposed in literature. It can be easily verified that these algorithms are much more complex than 3SOME and are enormously more computationally expensive, in terms of both computational overhead (i.e. computational procedures excluding the fitness evaluations) and memory employment. Finally it must be remarked that even though the Holm procedure does not allow us to fix performance-based hierarchy among 3SOME, DEGL, DEahcSPX, and SADE, 3SOME obtained the best performance for most of the test problems and, in some cases (see e.g. Fig. \ref{fig:perf3}), its application leads to much better results than the other algorithms. 

In order to highlight the simplicity features of 3SOME, the complexity of all the algorithms considered in this study is reported in Table \ref{complexity}. For each algorithm, besides a brief description of the main algorithmic features, the memory employment and computational overhead are shown. The memory employment is expressed in terms of minimum amount of memory slots required by the algorithm, where a memory slot is the memory space occupied by a vector having $n$ components, $n$ being the dimensionality of the problem. The memory requirement for scalar variables has been neglected for all the algorithms. The computational overhead has been calculated as the average run time over $30$ runs, for a given fitness landscape, $f_1$,  and a single core machine, required by each algorithm to perform $10000$ fitness evaluations. The machine  used to computer the overhead is a PC Intel Core 2 Duo 2.4 GHz with 4gb RAM employing GNU/Linux Ubuntu 10.04 and are obtained my means matlab 7.9.0.529 (R2009b).

\begin{table}
\caption{Algorithmic complexity of the algorithms} \label{complexity}
\begin{center} 
\begin{scriptsize}
\begin{tabular}{ c | c | c | c }
\hline \hline
Algorithm & Features & Computational Overhead [s] & Memory slots \\
\hline
EDA$_{mvg}$ & EDA structure & 2.7198 & $N_p(1+2\tau)+1$ \\ 
 & moving average system & & \\ \hline
FrankensteinPSO & PSO structure & 9.391 & $2N_p+neighborhood$ \\ 
 & adaptive inertia weight & & \\ 
 & reduced neighborhood & & \\ 
 & velocity restritcion & & \\ \hline  
2OptDE & DE structure & 3.4035 & $N_p+1$ \\ 
 & (novel mutation) & & \\ \hline  
DEcDE & compact DE based structure & 10.319 & $4$ \\ 
 & trigonometric mutation & & \\
 & stochastic perturbation & & \\ \hline  
(1+1)-CMA-ES & single-solution ES structure & 3.2706 & $n+2$ \\ 
 & covariance matrix driven search & & \\ \hline  
RCMA & GA structure & 5.9658 & $N_p+n+6$ \\
 & Solis-Wets Local Search & & \\ \hline  
DEGL & DE structure & 4.5467 & $N_p+1$ \\ 
 & index based neighborhood & & \\ \hline  
DEahcSPX & DE structure & 3.5168 & $N_p+n_p+2$ \\ 
 & Simplex Local Search & & \\ \hline  
SADE & DE structure & 4.8961 & $N_p+1+archive$ \\ 
 & multiple mutation strategies & & \\ 
 & self adaptive parameters & & \\ \hline  
3SOME & single-solution structure & 1.5246 & $3$ \\ 
 & 3 sequential operators & & \\ 
 & trial and arror coordination & & \\ \hline \hline
\end{tabular} 
\end{scriptsize}
\end{center} 
\end{table}        

\subsection{Bottom-up design of the algorithm: an experimental proof}

\begin{figure}
\centering
{\includegraphics[width=.8\linewidth]{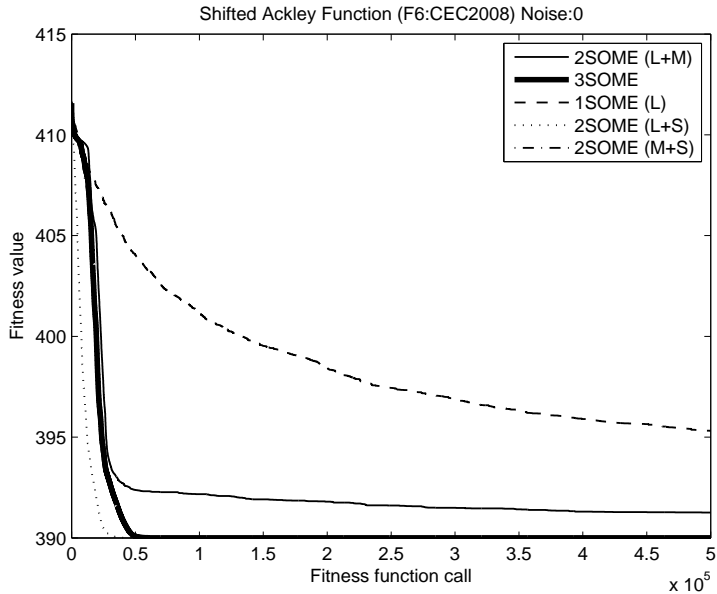}}
\caption{Performance trends of 1SOME, 2SOMEs, and 3SOME for $f_{29}$}\label{f:3SOMEvs2SOME}
\end{figure}

In order to show how the 3SOME algorithm has been designed and highlight the effectiveness of each component and their interaction, 3SOME performance has been compared with that of the stand alone long distance exploration, here indicated with 1SOME, and with the three possible combinations of algorithms composed of two models, here indicated with 2SOME. More specifically, 2SOME(L+M) employs only  long and middle distance exploration,  2SOME(L+S) employs only long and short distance exploration, and  2SOME(M+S) employs only middle and short distance exploration.  For the three versions of 2SOMEs, the operators are coordinated in the same way as in the 3SOME.  In this way, the role of each module is highlighted. These five algorithms have been run on the entire CEC 2008 setup for $n=100$, i.e. the functions $f_{24}-f_{30}$ in Section \ref{s:numres}. Numerical results of this experiment are displayed in Tables \ref{3SOMEvs2SOME} and \ref{HolmSOME},  and Fig. \ref{f:3SOMEvs2SOME}.


\begin{sidewaystable}
\caption{Numerical results of the 3SOME against 2SOMEs and 1SOME} \label{3SOMEvs2SOME}
\begin{center} 
\begin{scriptsize}
\begin{tabular}{c|c|c||c|c||c|c||c|c||c} 
\hline 
\hline  
\multicolumn{1}{l|}{Test Problem} & 1SOME & W& 2SOME(L+M) & W & 2SOME(L+S) & W & 2SOME(M+S) & W & 3SOME \\ \hline  
\multicolumn{ 1}{c|}{$f_{24}$}  & 9.345e+02 $\pm$ 2.04e+02 & + & 1.810e+01 $\pm$ 3.83e+00 & + & 9.663e-13 $\pm$ 1.19e-13 & = & 9.985e-13 $\pm$ 1.78e-13 & = & \bf{9.900e-13} $\pm$ \bf{1.70e-13} \\ \hline  
\multicolumn{ 1}{c|}{$f_{25}$}  & 4.153e+01 $\pm$ 3.69e+00 & + & 1.792e+00 $\pm$ 2.18e-01 & + & 7.296e-09 $\pm$ 2.03e-09 & = & 6.814e-01 $\pm$ 8.80e-01 & + & \bf{2.879e-09} $\pm$ \bf{1.11e-08} \\ \hline  
\multicolumn{ 1}{c|}{$f_{26}$}  & 4.463e+06 $\pm$ 1.39e+06 & + & 6.642e+03 $\pm$ 5.53e+03 & + & 3.170e+02 $\pm$ 4.63e+02 & = & 3.406e+02 $\pm$ 3.86e+02 & = & \bf{1.432e+02} $\pm$\bf{1.70e+02} \\ \hline  
\multicolumn{ 1}{c|}{$f_{27}$}  & 1.009e+02 $\pm$ 1.38e+01 & + & 1.497e+01 $\pm$ 4.16e+00 & + & 1.185e-12 $\pm$ 1.63e-13 & = & 1.197e-12 $\pm$ 1.64e-13 & = & \bf{1.132e-12} $\pm$ \bf{1.86e-13} \\ \hline  
\multicolumn{ 1}{c|}{$f_{28}$}  & 1.043e+01 $\pm$ 1.43e+00 & + & 1.175e+00 $\pm$ 2.92e-02 & + & 1.027e-02 $\pm$ 1.48e-02 & = & 2.670e-03 $\pm$ 4.53e-03 & = & \bf{2.978e-03} $\pm$ \bf{4.71e-03} \\ \hline  
\multicolumn{ 1}{c|}{$f_{29}$}  & 5.315e+00 $\pm$ 2.43e-01 & + & 1.262e+00 $\pm$ 1.54e-01 & + & 3.975e-12 $\pm$ 1.70e-13 & + & 2.260e-12 $\pm$ 2.20e-13 & = & \bf{2.141e-12} $\pm$ \bf{2.44e-13} \\ \hline  
\multicolumn{ 1}{c|}{$f_{30}$}  & -1.440e+03 $\pm$ 8.98e+00 & + & -1.486e+03 $\pm$ 1.04e+01 & = & -1.477e+03 $\pm$ 1.42e+01 & + & -1.485e+03 $\pm$ 1.55e+01 & = & \bf{-1.488e+03} $\pm$ \bf{1.83e+01} \\ \hline  \hline  
\end{tabular} 
\end{scriptsize}
\end{center} 
\end{sidewaystable} 

\begin{table}
\caption{Holm procedure for 1SOME, 2SOMEs, and 3SOME} \label{HolmSOME}
\begin{center} 
\begin{scriptsize}
\begin{tabular}{ c c c c c c } 
${N_A-j}$ & Algorithm & ${z_{N_A}-j}$ & ${p_{N_A}-j}$ & $\delta/(N_A-j)$ & Hypothesis \\ \hline \hline 
4 & 1SOME  & 0.00e+00 & 0.00e+00 & 1.25e-02 & Rejected \\ \hline  
3 & 2SOME(L+M) & -2.03e+00 & 2.13e-02 & 1.67e-02 & Accepted \\ \hline  
2 & 2SOME(L+S) & -6.76e-01 & 2.49e-01 & 2.50e-02 & Accepted \\ \hline  
1 & 2SOME(M+S) & -6.76e-01 & 2.49e-01 & 5.00e-02 & Accepted \\ \hline  
\end{tabular} 
\end{scriptsize}
\end{center} 
\end{table} 

Numerical results self-explicatively show the core philosophy of the bottom-up procedure. The starting point of our algorithmic design is not a successful algorithm but a \emph{tabula rasa} (blank slate). The first step of our design is to use a simple search operator which stochastically perturbs a solution within the entire decision space in order to detect new promising solutions. A simple stochastic global search (1SOME) improves upon an initial solution but its application  is not very effective as these improvements tend to occur excessively slowly. Thus, the alternate combination of global and local search (2SOME) appears to promisingly improve upon the performance of the 1SOME algorithm.  The coordination amongst the two 2SOME(L+M) components is trivial and natural. Middle distance exploration attempts to focus the search in the promising area detected by the long distance exploration. Thus, as soon as  the long distance exploration move detects a promising the solution, the search is focused around it by means of middle distance exploration which is continued until its application leads to benefit. When the short distance exploration applied directly after the long distance exploration, 2SOME(L+S), in some cases the algorithm fails, while in other cases achieves a good performance. This is due to the fact that the long distance exploration can, by itself detect a promising basin of attraction. The implicit use of the gradient of the third component can then quickly improve upon the solution. In the case of 2SOME(M+S) the result can be good under the condition that the initial sampling is done in the neighborhood of the optimal solution of that it is contained within the box whose side is $\delta$. All in all, although its superiority is not clearly confirmed by the Holm procedure, the complete sequence of modules (3SOME) appears to offer the best performance as it systematically seems to outperform the other algorithms under study.  Fig. \ref{f:3SOMEvs2SOME}  shows how each component is fundamental for the success in terms of performance.       

\subsection{Application: Digital IIR filter design}

In signal processing and digital control, it often happens that an optimization problem must be solved, see e.g. \cite{bib:DDITM}, \cite{bib:Karaboga2004}, \cite{bib:Karaboga2006} and\cite{bib:Tirronen2009}. In this subsection we test the performance of 3SOME with respect to the other nine algorithms under examination for the solution of a typical engineering problem, i.e. the design of a digital Infinite Impulse Response (IIR) filter, see \cite{bib:Karaboga2005}.  This problem consists of detecting the set of parameter which allows the identification of a plant. In other words, the IIR filter is an object which is supposed to simulate the behaviour of an unknown plant and its design means the minimization of a function which depends on the error between the behaviour of the plant (the reality) and the filter response (the model).  More specifically, the input-output relationship of a IIR filter, related to the generic sample $k$, is governed by the following formula:

\begin{equation}
y\left(k\right)=\sum_{i=1}^{M}b_i y\left(k-i\right)+\sum_{i=0}^{L}a_i u\left(k-i\right)
\end{equation}

\noindent where $u\left(k\right)$ and $y\left(k\right)$ are input and output, respectively, and $ M (\geq L)$ is the filter order. For each $i$, the constant values $a_i$ and $b_i$, define the problem. In other words, the general transfer function of the IIR filter can be written in the following way:
\begin{equation}
\frac{A\left(z\right)}{B\left(z\right)}=\frac{\sum_{i=0}^{L}a_iz^{-i}}{1+\sum_{i=1}^{M}b_iz^{-i}}.
\end{equation}

The IIR filter design consists of detecting that set of coefficients $a_i$ and $b_i$ which minimize the difference, in terms of behaviour, between model and reality. Thus, if we indicate with:

\begin{equation}
x=\left[a_0,a_1,\ldots, a_L, b_1, b_2,\ldots b_M\right],
\end{equation} 

we are willing to minimize the Mean Absolute Error, i.e. the following error function:

\begin{equation}\label{Jx}
J\left(x\right)=\frac{1}{N}\sum_{i=1}^{N}\left|d\left( k\right )- y\left( k\right )\right|
\end{equation} 

\noindent where $d\left( k\right )$ and $y\left( k\right )$ are respectively the measured actual plant response (which we want to reproduce) and the filter response of the $k^{th}$ sample, respectively, and $N$ is the total number of samples used to tune the filter.  The output of the actual plant is in general a non-linear function depending on the input $u\left( k\right )$.


In a real-world application, the transfer function of the plant would be unknown and the optimization process would lead towards the determination of a set of coefficients in $x$. In the simulated case, the transfer function is known but it is supposed unknown as the filter should be able to reconstruct the output of the plant.  The following transfer function simulates the plant in \cite{bib:Chen2002} :
\begin{equation}
H_P\left[z ^{-1} \right]= \frac{\alpha \left(z\right)} {\beta  \left(z\right)}
\end{equation} 

\noindent where 
\begin{eqnarray*}
\alpha\left(z\right)=z^{-1}-0.4 z ^{-2}+0.08z ^{-3}-0.032 z ^{-4} + 0.0816 z^{-5}+ \\
+ 0.0326 z^{-6} + 0.0288 z^{-7} - 0.0115 z^{-8} + 0.1296 z^{-9} - 0.0518 z^{-10} 
\end{eqnarray*}

\noindent and
\begin{eqnarray*}
\beta\left(z\right)={1 + 1.08 z^{-2} + 0.8726 z^{-4} + 0.6227 z^{-6} + 0.4694 z^{-8} + 0.1266 z^{-10}}.
\end{eqnarray*}

\begin{figure}
\centering
\subfigure[Output signal of plant and IIR filter after 3SOME optimization\label{output}]{\includegraphics[width=.49\linewidth]{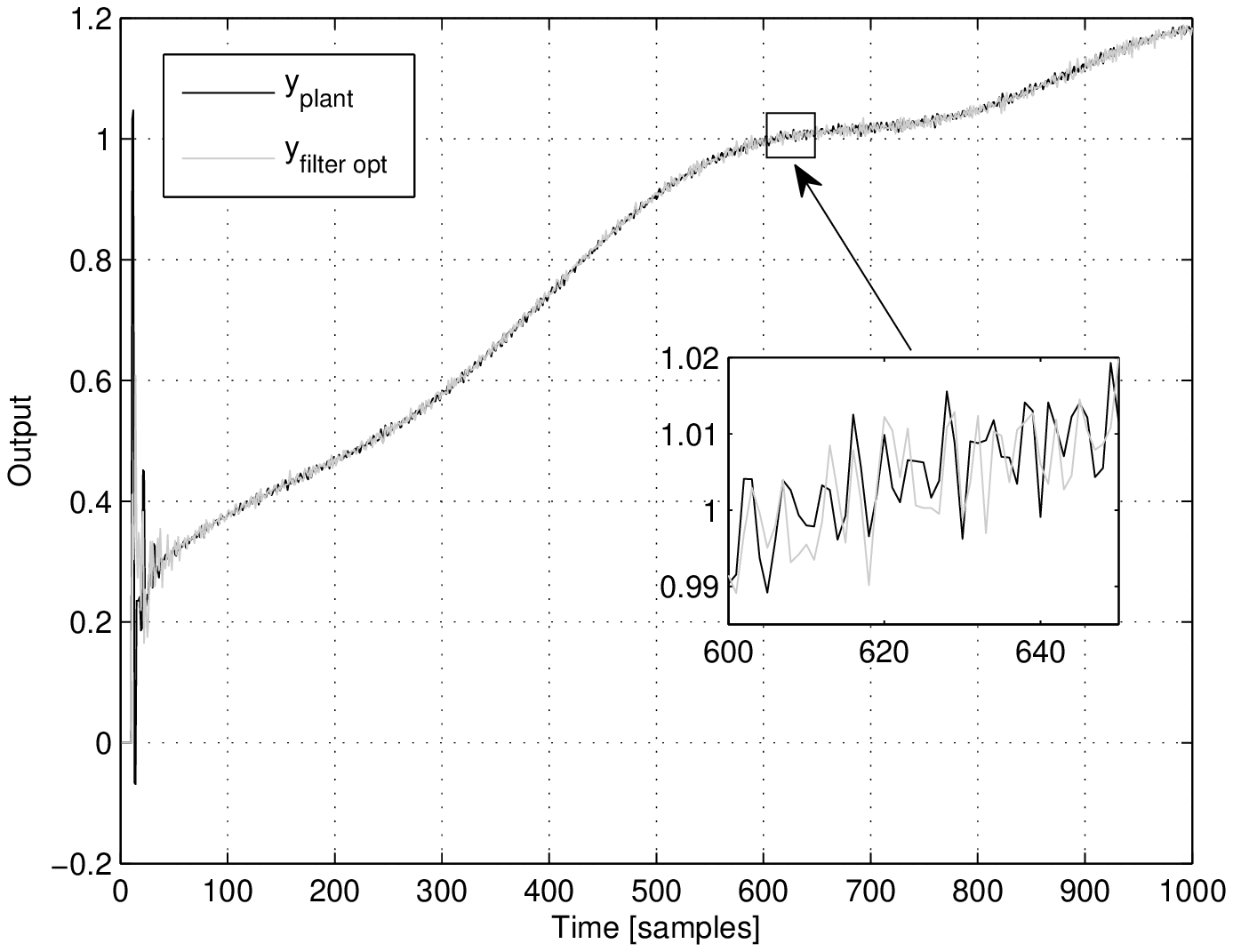}}
\subfigure[Pole-zero map referring to the plant and the IIR filter after 3SOME optimization\label{polezero}]{\includegraphics[width=.49\linewidth]{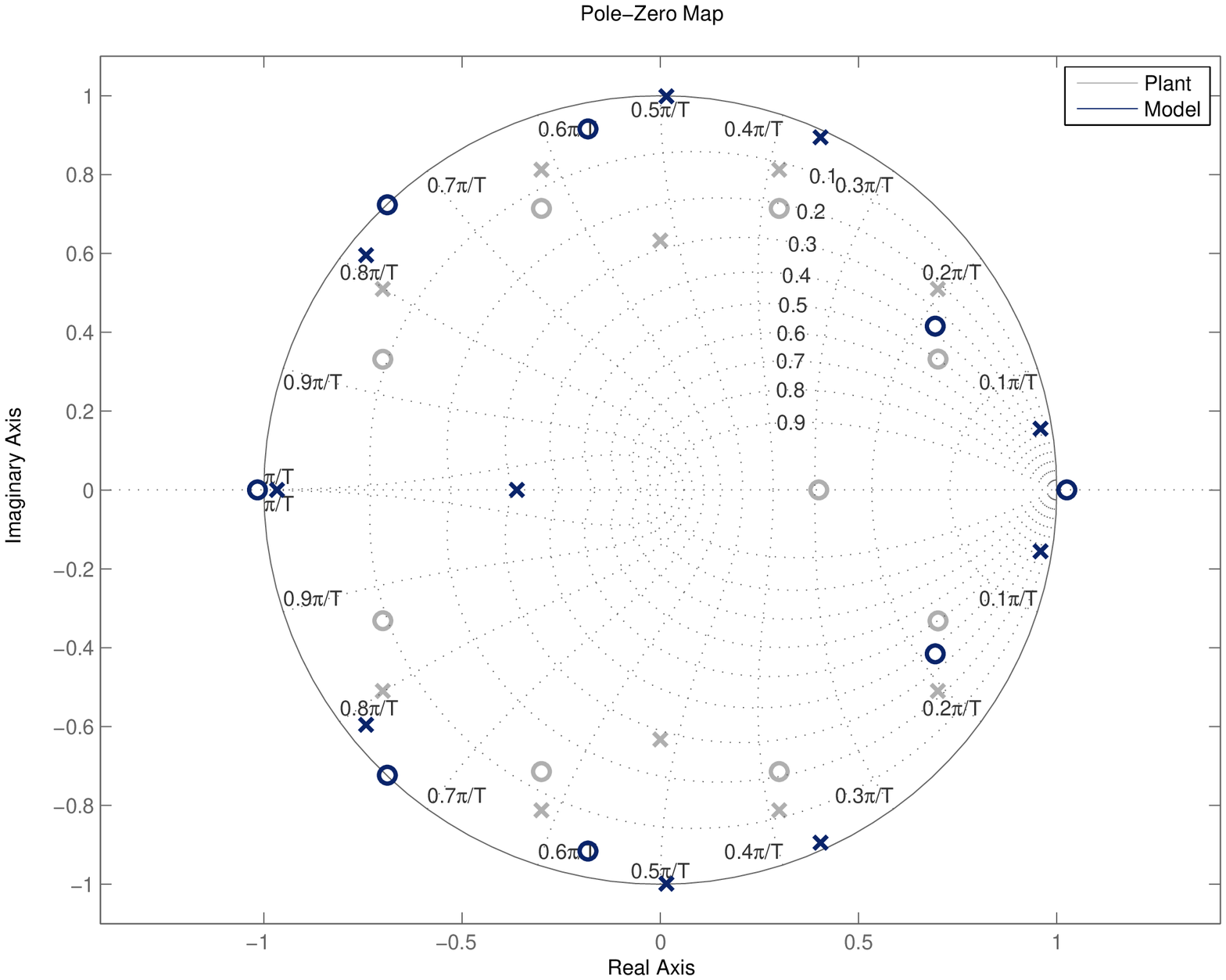}}
\caption{3SOME optimal solution in IIR filter design}
\end{figure}

In order to identify this plant, we used an IIR filter of the same order to the plant, i.e. a filter whose transfer function is of the following form:

 \begin{equation}
H_F\left[z ^{-1} \right]= \frac{a_0+a_1 z ^{-1}+\cdots a_i z ^{-i}+\cdots +a_{10} z ^{-10}}{1+b_1 z ^{-1}+\cdots + b_i z ^{-i}+ \cdots +b_{10} z ^{-10}},
\end{equation} 

\noindent and selected a number of samples $N=1000$. The system period is $T=0.001$ s and, in order to train the filter, we fed the plant with the following input signal:

\begin{equation}
u(k) = 1 + 5 sin\left(0.5\pi kT\right) + 0.25 sin(4 \pi k T + \phi) + 0.01 rand\left(0,1\right)
\end{equation} 

where $\phi=\frac{\pi}{3}$ and $rand\left(0,1\right)$ is a uniformly generated random number between $0$ and $1$. The minimization of $J\left(x\right)$ is performed in the decision space $D=\left[0,1\right]^{21}$ and a mechanism which heavily penalizes unstable solutions has been implemented. In order to recognize whether or not a candidate solution is unstable, each time a solution is generated, the position of the poles is checked that is within the unit circle (see Fig. \ref{polezero}). If the candidate solution turns out to be unstable, an infinite fitness value is assigned to it, otherwise the fitness value is computed according to the formula (\ref{Jx}).

The ten algorithms under considerations have been run for this application problem. For each algorithm, $30$ independent runs have been performed. Each run was allowed for $10000$ fitness evaluations. Table \ref{numfilter} shows the numerical results of this test.

\begin{table}
\caption{Optimization results for the IIR filter design problem} \label{numfilter}
\begin{center} 
\begin{tiny}
\begin{tabular}{c|c|c|c|c|c} 
\hline 
\hline  
& SADE & DEGL & FrankensteinPSO & DEahcSPX & DEcDE \\ \hline 
Mean & 4.6788e-03 & 4.5279e-02 & 7.7009e-03 & 1.3375e-02 &  1.2480e-02 \\ \hline 
Std. Dev. & 1.26e-03 & 3.75e-02 & 7.03e-04 & 3.23e-03 &  1.34e-03\\ \hline 
Wilcoxon & - & + & - & = &= \\ \hline \hline 
  EDA$_{mvg}$ & RCMA & (1+1)-CMA-ES & 2OptDE & 3SOME \\ \hline  
9.8372e-03 & 7.5292e-03 & 1.2567e+00 & 2.8494e-01 & 1.6743e-02 \\ \hline  
  1.31e-03 & 7.45e-04 & 3.53e+00 & 2.39e-01 & 1.01e-02 \\ \hline 
   = & - & + & + & \\ \hline  \hline
\end{tabular} 
\end{tiny}
\end{center} 
\end{table} 


Numerical results show that for this application problems 2OptDE and (1+1)-CMA-ES have a worse performance compared to the other algorithms. The remaining eight algorithms are quite similar, at least from an engineering viewpoint. The proposed 3SOME displays a respectable performance and is competitive with the other more complex approaches. 

In order to better explain the physical meaning of the application the response $y$ of the plant and of the IIR filter after the  optimization (performed by 3SOME) is plotted in Fig. \ref{output}.  It can be observed that the filter approximates satisfactorily the signal of the plant. Fig. \ref{polezero} displays the pole-zero map related to the plant transfer function and to the IIR filter after optimization performed by 3SOME. It must be observed that the detected solution guarantees a stable behaviour of the system (all the poles are contained within the unit circle).

\section{Conclusion}\label{s:conc}
This paper shows that, in MC,  a very simple algorithm, if properly designed, can outperform much more complex and computationally expensive approaches.  The proposed 3SOME algorithm is composed of  three exploration operators, characterized by different pivot rules and exploratory radius, and are coordinated by means of trial and error mechanism. Numerical results show that 3SOME is very competitive, for the problems and dimensionality values reported,  with modern algorithms generated by modifying and combining already existing structures. Thus, the proposed algorithm is suitable for implementation in those contexts where hardware limitations impose a limited hardware and overhead usage. Most importantly, this paper aims at highlighting the importance of the bottom-up approach in algorithmic design. This allows a better understanding of each operator involved and helps to maintain a clean and elegant algorithmic design. In other words, since algorithms recently proposed in literature resulted to be overwhelmingly complex with respect to their performance, a greater attention should be likely paid to the algorithmic design phase which should be accompanied by an algorithmic philosophy, i.e. the presence of each element should be intuitively understood and justified. In this light, the law of parsimony by William Ockham appears to us as an appropriate philosophical guidance and source of inspiration.

\section*{Appendix: Test Problems}\label{test_desc}
\noindent The following test problems have been considered in this study.
\\ \\
\noindent $f_1$ Shifted sphere function: $f_1$ from \cite{bib:Suganthan2005} with $n=30$.
\begin{equation}\label{f1}
   f_1\left(x\right)=\sum_{i=1}^{n}{z_i^2}
\end{equation}
\noindent where $z=x-o$ and the shifted optimum $o=\left[o_1,o_2,\ldots,o_n\right]$. 
Decision space $D=\left[-100,100\right]^n$. Properties: Unimodal, Shifted, Separable, Scalable.
\\ \\
\noindent $f_2$ Shifted Schwefel's Problem 1.2: $f_2$ from \cite{bib:Suganthan2005} with $n=30$.
\begin{equation}\label{f2}
f_2\left(x\right)=\sum_{i=1}^{n}\left(\sum_{j=1}^{i}z_j\right)^2
\end{equation}
\noindent where $z=x-o$ and the shifted optimum $o=\left[o_1,o_2,\ldots,o_n\right]$. 
Decision space $D=\left[-100,100\right]^n$. Properties: Unimodal, Shifted, Non-separable, Scalable.
\\ \\
\noindent $f_3$ Rosenbrock's function: $f_3$ from \cite{bib:Qin2009} with $n=30$.
\begin{equation}\label{f3}
f_3\left(x\right)=\sum_{i=1}^{n-1}\left(100 \left(x_{i+1} -
x_i^2\right)^2+\left(1-x_i\right)^2\right).
\end{equation}
Decision space $D=\left[-100,100\right]^n$. Properties: Multi-modal, Non-separable, Scalable.
\\ \\
\noindent $f_4$ Shifted Ackley's function: $f_5$ from \cite{bib:Qin2009} with $n=30$.
\begin{equation}\label{f4}
f_4\left(x\right)=-20 e^{-0.2 \sqrt{1/n\sum_{i=1}^{n}z_i^2}} - e^{\left(1/n\right) 
\sum_{i=1}^{n}{\cos(2\pi z_i)}} + 20 + e
\end{equation}
\noindent where $z=x-o$ and the shifted optimum $o=\left[o_1,o_2,\ldots,o_n\right]$. 
Decision space $D=\left[-32,32\right]^n$. Properties: Multi-modal, Shifted, Non-separable, Scalable.
\\ \\
\noindent $f_5$ Shifted rotated Ackley's function: $f_6$ from \cite{bib:Qin2009} with $n=30$.
The formula used is the same as equation \ref{f4}, where $z=M(x-o)$, condition number of matrix $M$ being 
equal to $1$, and the shifted optimum $o=\left[o_1,o_2,\ldots,o_n\right]$. 
Decision space $D=\left[-32,32\right]^n$. Properties: Multi-modal, Rotated, Shifted, Non-separable, Scalable.
\\ \\
\noindent $f_6$ Shifted Griewank's function: $f_7$ from \cite{bib:Qin2009} with $n=30$.
\begin{equation}\label{f6}
f_6\left(x\right)=\sum_{i=1}^{n}\frac{z_i^2}{4000}-\prod_{i=1}^{n}\cos\frac{z_{i}}{\sqrt{i}}+1
\end{equation}
\noindent where $z=x-o$ and the shifted optimum $o=\left[o_1,o_2,\ldots,o_n\right]$. 
Decision space $D=\left[-600,600\right]^n$. Properties: Multi-modal. Shifted, Non-separable, Scalable.
\\ \\
\noindent $f_7$ Shifted rotated Griewank's function: $f_8$ from \cite{bib:Qin2009} with $n=30$.
The formula used is the same as equation \ref{f6}, where $z=M(x-o)$, condition number of matrix $M$ being
equal to $3$, and the shifted optimum $o=\left[o_1,o_2,\ldots,o_n\right]$. 
Decision space $D=\left[-600,600\right]^n$. Properties: Multi-modal, Rotated, Shifted, Non-separable, Scalable.
\\ \\
\noindent $f_8$ Shifted Rastrigin's function: $f_9$ from \cite{bib:Suganthan2005} with $n=30$.
\begin{equation}\label{f8}
f_8\left(x\right)=10n+\sum_{i=1}^{n}\left(z_i^2-10\cos{2\pi z_i}\right)
\end{equation}
\noindent where $z=x-o$ and the shifted optimum $o=\left[o_1,o_2,\ldots,o_n\right]$. 
Decision space $D=\left[-5,5\right]^n$. Properties: Multi-modal, Shifted, Separable, Scalable, huge number of local optima.
\\ \\
\noindent $f_9$ Shifted rotated Rastrigin's function: $f_{10}$ from \cite{bib:Suganthan2005} with $n=30$.
The formula used is the same as equation \ref{f8}, where $z=M(x-o)$, condition number of matrix $M$ being
equal to $3$, and the shifted optimum $o=\left[o_1,o_2,\ldots,o_n\right]$. 
Decision space $D=\left[-5,5\right]^n$. Properties: Multi-modal, Shifted, Rotated, Separable, Scalable, huge number of local optima.
\\ \\
\noindent $f_{10}$ Shifted non continuous Rastrigin's function: $f_{11}$ from \cite{bib:Qin2009} with $n=30$.
\begin{equation}\label{f10}
f_{10}\left(x\right)=10n+\sum_{i=1}^{n}\left(y_i^2-10\cos{2\pi y_i}\right)
\end{equation}
\begin{equation}
y_i = \left\{
\begin{array}{l l}
 z_i & \quad \mbox{if } \left| z_i \right| < 1/2 \\
 round(2z_i)/2 & \quad \mbox{if } \left| z_i \right| \geq 1/2\\ \end{array} \right.
\end{equation}
\noindent where $z=x-o$ and the shifted optimum $o=\left[o_1,o_2,\ldots,o_n\right]$. 
Decision space $D=\left[-500,500\right]^n$. Properties: Multi-modal, Shifted, Rotated, Separable, Scalable, huge number of local optima.
\\ \\
\noindent $f_{11}$ Schwefel's function: $f_{12}$ from \cite{bib:Qin2009} with $n=30$.
\begin{equation}\label{f11}
f_{11}\left(x\right)=418.9829n + \sum_{i=1}^{n}\left(-x_i \sin{\sqrt{\left|x_i\right|}}\right).
\end{equation}
Decision space $D=\left[-500,500\right]^n$. Properties: Multi-modal, Separable, Scalable.
\\ \\
\noindent $f_{12}$ Schwefel Problem 2.22: $f_2$ from \cite{bib:Vestersrtom2004} with $n=10$.
\begin{equation}\label{f12}
f_{12}\left(x\right)=\sum_{i=1}^{n}\left|x_i\right| + \prod_{i=1}^{n}\left|x_i\right|.
\end{equation}
Decision space $D=\left[-10,10\right]^{n}$. Properties: Unimodal, Separable, Scalable.
\\ \\
\noindent $f_{13}$ Schwefel Problem 2.21: $f_4$ from \cite{bib:Vestersrtom2004} with $n=10$.
\begin{equation}\label{f13}
f_{13}\left(x\right)=\max_{i}\left|x_i\right|.
\end{equation}
Decision space $D=\left[-100,100\right]^{n}$. Properties: Unimodal, Non-separable, Scalable.
\\ \\
\noindent $f_{14}$ Generalized penalized function 1: $f_{12}$ from \cite{bib:Vestersrtom2004} with $n=10$.
\begin{equation}\label{f14}
\begin{array}{c}
f_{14}\left(x\right)=\frac{\pi}{n} \left\lbrace 10 \sin^2{\pi y_1} + 
\sum_{i=1}^{n}\left( (y_i-1)^2 \left(1+10\sin^2\pi y_i \right) \right) + 
\left( y_n-1 \right)^2 \right\rbrace  \\ + \sum_{i=1}^{n} u(x_i,10,100,4)
\end{array}
\end{equation}
\noindent where,
\begin{equation}
 y_i=1+\frac{1}{4}(x_i+1)
\end{equation}
and,
\begin{equation}
 u(x,a,k,m)=\left\{
\begin{array}{l l}
 k(x_i-a)^m &     \quad \mbox{if } x_i > a \\
 0        \quad \mbox{if } \left|x_i\right| \leq a \\
 k(-x_i-a)^m & \quad \mbox{if } x_i < -a \\
\end{array} \right.
\end{equation}
Decision space $D=\left[-50,50\right]^{n}$. Properties: Multi-modal, Separable, Scalable.
\\ \\
\noindent $f_{15}$ Generalized penalized function 2: $f_{13}$ from \cite{bib:Vestersrtom2004} with $n=10$.
\begin{equation}\label{f15}
\begin{array}{c}
f_{15}\left(x\right)= \\ \frac{1}{10} \left\lbrace \sin^23\pi x_1 +  \sum_{i=1}^{n-1}\left( (x_i-1)^2 \left(1+\sin^23\pi x_{i+1}\right) \right)+ \left(x_n -1\right)\left(1+\sin2\pi x_n \right)^2 \right\rbrace + \\ \sum_{i=1}^{n} u(x_i,5,100,4)
\end{array}
\end{equation}
Decision space $D=\left[-50,50\right]^{n}$. Properties: Multi-modal, Separable, Scalable.
\\ \\
\noindent $f_{16}$ Schwefel's Problem 2.6 with Global Optimum on Bounds: $f_5$ from \cite{bib:Suganthan2005} with $n=30$.
Decision space $D=\left[-100,100\right]^n$. Properties: Unimodal, Non-separable, Scalable.
\\ \\
\noindent $f_{17}$ Shifted Rotated Weierstrass Function: $f_{11}$ from \cite{bib:Suganthan2005} with $n=30$.
Decision space $D=\left[-0.5,0.5\right]^n$. Properties: Multi-modal, Shifted, Rotated, Non-separable, Scalable, 
Continuous but differentiable only on a set of points.
\\ \\
\noindent $f_{18}$ Schwefel's Problem 2.13: $f_{12}$ from \cite{bib:Suganthan2005} with $n=30$.
Decision space $D=\left[-\pi,\pi\right]^n$. Properties: Properties: Multi-modal, Shifted, Non-separable, Scalable.
\\ \\
\noindent $f_{19}$ Shifted rotated Rastrigin's function: the same as $f_9$, with same bounds and $n=50$.
Properties: Multi-modal, Shifted, Rotated, Separable, Scalable, huge number of local optima.
\\ \\
\noindent $f_{20}$ Michalewicz's function: from \cite{bib:mich} with $n=50$.
\begin{equation}\label{f20}
f_{20}\left(x\right)=-\sum_{i=1}^{n}\sin\left(x_i\right){\left[\sin{\left(\frac{ix_i^2}{\pi}\right)}\right]}^{2m}
\end{equation}
where $m=10$. Decision space $D=\left[0,\pi\right]^{n}$. Properties: Multi-modal, Separable, Scalable.
\\ \\
\noindent $f_{21}$ Schwefel's function (see equation \ref{f11}), with same bounds and $n=50$. 
Properties: Multi-modal, Separable, Scalable.
\\ \\
\noindent $f_{22}$ Michalewicz's function (see equation \ref{f20}), with same bounds and $n=100$. 
Properties: Multi-modal, Separable, Scalable.
\\ \\
\noindent $f_{23}$ Schwefel's function (see equation \ref{f11}), with same bounds and $n=100$. 
Properties: Multi-modal, Separable, Scalable.
\\ \\
\noindent $f_{24}$ Shifted sphere function: $f_1$ from \cite{bib:Tang2007} (see equation \ref{f1}), with 
same bounds and $n=100$. Properties: Unimodal, Shifted, Separable, Scalable.
\\ \\
\noindent $f_{25}$ Shifted Schwefel Problem 2.21: $f_2$ from \cite{bib:Tang2007} with $n=100$.
\begin{equation}\label{f13}
f_{13}\left(x\right)=\max_{i}\left|z_i\right|
\end{equation} 
\noindent where $z=x-o$ and the shifted optimum $o=\left[o_1,o_2,\ldots,o_n\right]$. 
Decision space $D=\left[-100,100\right]^{n}$. Properties: Unimodal, Shifted, Non-separable, Scalable.
\\ \\
\noindent $f_{26}$ Shifted Rosenbrock's Function: $f_3$ from \cite{bib:Tang2007} 
$n=100$
Properties: Multi-modal, Shifted, Non-separable, Scalable.
\\ \\
\noindent $f_{27}$ Shifted Rastrigin's Function: $f_4$ from \cite{bib:Tang2007} (see equation \ref{f8}), with
same bounds and $n=100$. Properties: Multi-modal, Shifted, Separable, Scalable, huge number is huge of local optima.
\\ \\
\noindent $f_{28}$ Shifted Griewank's Function: $f_5$ from \cite{bib:Tang2007} (see equation \ref{f6}), with
same bounds and $n=100$. Properties: Multi-modal, Shifted, Non-separable, Scalable.
\\ \\
\noindent $f_{29}$ Shifted Ackley's Function: $f_6$ from \cite{bib:Tang2007} (see equation \ref{f4}), with 
same bounds and $n=100$. Properties: Multi-modal, Shifted, Non-separable, Scalable.
\\ \\
\noindent $f_{30}$ FastFractal DoubleDip Function: $f_7$ from \cite{bib:Tang2007} with $n=100$.
\begin{equation}\label{f30}
f_{30}\left(x\right)=\sum_{i=1}^{D}fractal1D\left(x_i+twist\left(x_{(i\mathrm{mod}D)+1}\right)\right)
\end{equation}
\begin{equation}
twist\left(x\right)=4\left(y^4-2^3+y^2\right)
\end{equation}
\begin{equation}
fractal1D\left(x\right)\approx\sum_{k=1}^{3}\sum_{1}^{2^{k-1}}\sum_{1}^{ran2\left(o\right)}
doubledip\left(x,ran1\left(o\right),\frac{1}{2^{k-1}\left(2-ran1\left(o\right)\right)}\right)
\end{equation}
\begin{equation}
doubledip\left(x,c,s\right) = \left\{\begin{array}{l l}
 \left(-6144\left(x-c\right)^6-3088\left(x-c\right)^4-392\left(x-c\right)^2+1\right)s \mbox{, }\\ -0.5 < x < 0.5 \\
 0\mbox{, otherwise} \\ 
\end{array} \right.
\end{equation}
where $ran1(o)$ and $ran2(o)$ are, respectively, a double and an integer, pseudo-randomly chosen, with seed o, with 
equal probability from the interval $[0,1]$ and the set $\{0,1,2\}$.
Decision space $D=\left[-1,1\right]^{n}$. Properties: Multi-modal, Non-separable, Scalable.

\section*{Acknowledgements}
This work is supported by Academy of Finland,  Akatemiatutkija 130600, Algorithmic Design Issues in Memetic Computing and Tutkijatohtori 140487,  Algorithmic Design and Software Implementation: a Novel Optimization Platform. This research is supported also by University of Jyv{\"a}skyl{\"a}, Mobility Grants 2011. 
\bibliographystyle{elsarticle-harv}
\bibliography{biblio}

\end{document}